\documentclass[runningheads]{llncs}
\usepackage{graphicx}
\usepackage[misc]{ifsym}
\usepackage{booktabs}
\usepackage{graphicx}
\usepackage{pifont}
\usepackage{amsmath}
\usepackage{amssymb}
\usepackage{amsfonts}
\usepackage{bbm}
\usepackage{subcaption}
\usepackage[dvipsnames]{xcolor}
\usepackage{multirow}
\usepackage{url}

\begin{document}
\title{F\textsc{orecast}TKGQ\textsc{uestions}: A Benchmark for Temporal Question Answering and Forecasting over Temporal Knowledge Graphs}
\toctitle{F\textsc{orecast}TKGQ\textsc{uestions}: A Benchmark for Temporal Question Answering and Forecasting over Temporal Knowledge Graphs}

\author{Zifeng Ding\footnote[1]{Equal contribution.}\inst{1,2} \and Zongyue Li\inst{\star 1,3} \and Ruoxia Qi\inst{\star 1} \and Jingpei Wu\inst{1} \and Bailan He\inst{1,2} \and \\ Yunpu Ma\inst{1,2} \and Zhao Meng\inst{4} \and Shuo Chen\inst{1,2} \and  Ruotong Liao\inst{1,3} \and \\Zhen Han(\Letter)\inst{1} \and Volker Tresp(\Letter)\inst{1}}
\authorrunning{Ding et al.} 
%
\tocauthor{Zifeng Ding, Zongyue Li, Ruoxia Qi, Jingpei Wu, Bailan He, Yunpu Ma, Zhao Meng, Shuo Chen, Ruotong Liao, Zhen Han, Volker Tresp}
\institute{LMU Munich, Geschwister-Scholl-Platz 1, 80539 Munich, Germany
\and
Siemens AG, Otto-Hahn-Ring 6, 81739 Munich, Germany
\and Munich Center for Machine Learning (MCML), Munich, Germany
\and ETH Zürich, Rämistrasse 101, 8092 Zürich, Switzerland
\\
\email{\{zifeng.ding, ruoxia.qi, bailan.he, shuo.chen\}@campus.lmu.de, \\ \{zongyue.li, jingpei.wu\}@outlook.com, cognitive.yunpu@gmail.com, zhmeng@ethz.ch, liao@dbs.ifi.lmu.de, hanzhen02111@hotmail.com, Volker.Tresp@lmu.de}
}

\titlerunning{A Benchmark for Temporal QA and Forecasting over TKGs}
%
\maketitle              
\begin{abstract}
Question answering over temporal knowledge graphs (TKGQ\\A) has recently found increasing interest. 
Previous related work aims to develop QA systems that answer temporal questions based on the facts from a fixed time period, where a temporal knowledge graph (TKG) spanning this period can be fully used for answer inference. In real-world scenarios, however, it is also common that given the knowledge until now, we wish the TKGQA systems to answer the questions asking about the future. As humans constantly seek plans for the future, building forecasting TKGQA systems is important.
In this paper, we propose a novel task: forecasting TKGQA, and propose a coupled large-scale TKGQA benchmark dataset, i.e., F\textsc{orecast}TKGQ\textsc{uestions}. It includes three types of forecasting questions, i.e., entity prediction, yes-unknown, and fact reasoning questions. 
For every forecasting question, a timestamp is annotated and QA models can only have access to the TKG information before it for answer inference.
We find that previous TKGQA methods perform poorly on forecasting questions, and they are unable to answer yes-unknown and fact reasoning questions. To this end, we propose F\textsc{orecast}TKGQA, a TKGQA model that employs a TKG forecasting module for future inference.
Experimental results show that F\textsc{orecast}TKGQA performs well in answering forecasting questions.
\end{abstract}
\section{Introduction}
Knowledge graphs (KGs) model factual information by representing every fact with a triplet, i.e., $(s,r,o)$, where $s$, $o$, $r$, are the subject entity, the object entity, and the relation between $s$ and $o$, respectively. To adapt to the ever-evolving knowledge,
temporal knowledge graphs (TKGs) are introduced, where they additionally specify the time validity of every fact with a time constraint $t$ ($t$ is a timestamp), and represent each fact with a quadruple $(s,r,o,t)$.
Recently, TKG reasoning has drawn great attention. While a lot of methods 
focus on temporal knowledge graph completion (TKGC) where they predict missing facts at the observed timestamps, various recent methods 
pay more attention to forecasting the facts at unobserved future timestamps in TKGs.

Knowledge graph question answering (KGQA) is a task aiming to answer the natural language questions using a KG as the knowledge base (KB). KGQA requires QA models to extract answers from KGs, rather than retrieving or summarizing answers from the given text contexts. 
\cite{DBLP:conf/acl/SaxenaCT20} first introduces question answering over temporal knowledge graphs (TKGQA). It proposes a non-forecasting TKGQA dataset C\textsc{ron}Q\textsc{uestions} that takes a TKG as its underlying KB.  
Temporal reasoning techniques are required to answer these questions. Though \cite{DBLP:conf/acl/SaxenaCT20} manages to combine TKG reasoning with KGQA, it has limitations. Previous KGQA datasets, including C\textsc{ron}Q\textsc{uestions}, do not include yes-no and multiple-choice questions, while these two question types have been extensively studied in reading comprehension QA, e.g., \cite{DBLP:conf/acl/JinKKLMGR20}. Besides, the questions in C\textsc{ron}Q\textsc{uestions} are in a non-forecasting style, where all of them are based on the TKG facts that happen in a fixed time period, and an extensive TKG that is fully observable in this period can be used to infer the answers, making the answer inference less challenging. For example, the TKG facts from \textit{2003}, including (\textit{Stephen Robert Jordan}, \textit{member of sports team}, \textit{Manchester City}, \textit{2003}), are all observable to answer the question \textit{Which team was Stephen Robert Jordan part of in 2003?}. C\textsc{ron}Q\textsc{uestions} manages to bridge the gap between TKGC and KGQA, however, no previous work manages to combine TKG forecasting with KGQA, where only past TKG information can be used for answer inference.


In this work, we propose a novel task: forecasting question answering over temporal knowledge graphs (forecasting TKGQA), together with a coupled large-scale dataset, i.e., F\textsc{orecast}TKGQ\textsc{uestions}. We generate forecasting questions based on the Integrated Crisis Early Warning System (ICEWS) dataverse
\cite{DVN/28075_2015}, and label every question with a timestamp. To answer a forecasting question, QA models can only access the TKG information prior to the question timestamp. 
The contribution of our work is three-folded:
(1) We propose forecasting TKGQA, a novel task aiming to test the forecasting ability of TKGQA models. To the best of our knowledge, this is the first work binding TKG forecasting with temporal KGQA; (2) We propose a large-scale benchmark TKGQA dataset: F\textsc{orecast}TKGQ\textsc{uestions}. It contains three types of questions, i.e., entity prediction questions (EPQs), yes-unknown questions (YUQs), and fact reasoning questions (FRQs), where the last two types of questions have never been considered in previous KGQA datasets\footnote{YUQs are based on yes-no questions and FRQs are multiple-choice questions.}; (3) We propose F\textsc{orecast}TKGQA, a model aiming to solve forecasting TKGQA. It employs a TKG forecasting module and a pre-trained language model (LM) for answer inference. Experimental results show that it achieves great performance on forecasting questions.
\section{Preliminaries and Related Work}
\subsubsection{TKG Reasoning}
Let $\mathcal{E}$, $\mathcal{R}$ and $\mathcal{T}$ denote a finite set of entities, relations, and timestamps, respectively. A TKG $\mathcal{G}$ is defined as a finite set of TKG facts represented by quadruples, i.e., $\mathcal{G} = \{(s,r,o,t)|s,o \in \mathcal{E}, r \in \mathcal{R}, t\in \mathcal{T}\}$. We define the TKG forecasting task (also known as TKG extrapolation) as follows. Assume we have a query $(s_q,r_q,?,t_q)$ (or $(?,r_q,o_q,t_q)$) derived from a target quadruple $(s_q,r_q,o_q,t_q)$, and we denote all the ground-truth quadruples as $\mathcal{F}$. TKG forecasting aims to predict the missing entity in the query, given the observed \textbf{past} TKG facts $\mathcal{O} = \{(s_i, r_i, o_i, t_i)\in \mathcal{F}|t_i < t_q\}$. Such temporal restriction is not imposed in TKG completion (TKGC, also known as TKG interpolation), where the observed TKG facts from any timestamp, including $t_q$ and the timestamps after $t_q$, can be used for prediction. In recent years, there have been extensive works done for both TKGC \cite{DBLP:conf/iclr/LacroixOU20,DBLP:conf/kdd/JungJK21,ding2022a} and TKG forecasting \cite{DBLP:conf/emnlp/JinQJR20,DBLP:conf/emnlp/HanDMGT21,DBLP:conf/aaai/ZhuCFCZ21,DBLP:conf/iclr/HanCMT21,DBLP:conf/aaai/LiuMHJT22}. We give a more detailed discussion about the forecasting methods. RE-N\textsc{ET} \cite{DBLP:conf/emnlp/JinQJR20} employs an autoregressive architecture and models fact occurrence as a probability distribution conditioned on the temporal sequences of past related TKG information. TANGO \cite{DBLP:conf/emnlp/HanDMGT21} employs neural ordinary differential equations to model temporal dependencies among graph information of different timestamps. CyGNet \cite{DBLP:conf/aaai/ZhuCFCZ21} uses the copy-generation mechanism to extract hints from historical facts for forecasting. xERTE \cite{DBLP:conf/iclr/HanCMT21} constructs a historical fact-based subgraph and selects prediction answers from it. TLogic \cite{DBLP:conf/aaai/LiuMHJT22} is the first rule-based TKG forecasting method that learns temporal logical rules in TKGs and achieves superior results.

\subsubsection{Question Answering over KGs}Several datasets have been proposed for QA over non-temporal KGs, such as SimpleQuestions \cite{https://doi.org/10.48550/arxiv.1506.02075}, WebQuestionsSP \cite{DBLP:conf/acl/YihCHG15}, ComplexWebQuestions \cite{DBLP:conf/naacl/TalmorB18}, MetaQA \cite{DBLP:conf/aaai/ZhangDKSS18}, TempQuestions \cite{DBLP:conf/www/JiaARSW18}, and TimeQuestions \cite{DBLP:conf/cikm/JiaPRW21}. 
Among these datasets, only TempQuestions and TimeQuestions involve temporal questions that require temporal reasoning for answer inference, however, their associated KGs are non-temporal.
C\textsc{ron}Q\textsc{uestions} \cite{DBLP:conf/acl/SaxenaCT20} contains questions based on a time-evolving TKG, i.e., Wikidata \cite{DBLP:journals/cacm/VrandecicK14}. It is proposed for non-forecasting TKGQA. Two types of questions, i.e., entity prediction and time prediction questions, are included. To answer C\textsc{ron}Q\textsc{uestions}, Saxena et al. propose C\textsc{ron}KGQA that uses TKGC methods, along with pre-trained LMs, which shows great effectiveness. A line of methods has been proposed on top of C\textsc{ron}KGQA (TempoQR \cite{DBLP:conf/aaai/MavromatisSIAHG22}, TSQA \cite{DBLP:conf/acl/ShangW0022}, SubGTR \cite{DBLP:journals/kbs/ChenZLLK22}),
where they better distinguish question time scopes and reason over subgraphs. C\textsc{ron}Q\textsc{uestions} is proposed based on the idea of TKGC, and it does not support TKG forecasting and contains no forecasting question. One recent work, i.e., F\textsc{orecast}QA \cite{DBLP:conf/acl/JinKKLMGR20}, proposes a QA dataset fully consisting of forecasting questions. 
However, F\textsc{orecast}QA is not related to KGQA. In F\textsc{orecast}QA, answers to its questions are inferred from text contexts, while KGQA/TKGQA requires models to find the answers from the coupled KGs/TKGs without providing any additional text contexts. As a result, the methods designed for F\textsc{orecast}QA have no ability to address TKGQA. To this end, we propose F\textsc{orecast}TKGQ\textsc{uestions}, aiming to bridge the gap between TKG forecasting and KGQA. We compare F\textsc{orecast}TKGQ\textsc{uestions} with recent KGQA datasets in Table \ref{tab: dataset compare}.

\begin{table}[t]
    \caption{(a) KGQA dataset comparison. Statistics are taken from \cite{DBLP:conf/acl/SaxenaCT20} and \cite{DBLP:conf/cikm/JiaPRW21}. \textbf{T\%} denotes the portion of temporal questions. (b) F\textsc{orecast}TKGQ\textsc{uestions} statistics: number of questions of different types.}
    \begin{subtable}{.57\linewidth}
      \centering
        \caption{}\label{tab: dataset compare}
    \resizebox{0.98\columnwidth}{!}{
    \begin{tabular}{@{}c|c|c|c|c @{}}
        \textbf{Datasets} & \textbf{TKG} & \textbf{Forecast}&\textbf{T\%} & \textbf{\# Questions}\\

\hline
MetaQA  & \ding{55} & \ding{55}& 0\% & 400k\\
TempQuestions  & \ding{55} & \ding{55} & 100\% & 1271\\
TimeQuestions  & \ding{55} & \ding{55} & 100\% & 16k\\
C\textsc{ron}Q\textsc{uestions}  & \ding{51} & \ding{55} & 100\% & 410k\\
F\textsc{orecast}TKGQ\textsc{uestions}  & \ding{51} & \ding{51}& 100\% & 727k\\
    \end{tabular}
    }
    \end{subtable}%
    \begin{subtable}{.43\linewidth}
      \centering
        \caption{}\label{tab: dataset statistics}
    \resizebox{0.98\columnwidth}{!}{
    \begin{tabular}{@{}c|c|c|c @{}}
         & \textbf{Train} & \textbf{Valid} & \textbf{Test}\\

\hline
1-Hop Entity Prediction& 211,564 & 36,172 & 33,447 \\
2-Hop Entity Prediction & 85,088 & 12,266 & 10,765 \\
Yes-Unknown & 251,537 & 42,884 & 39,695 \\
Fact Reasoning & 3,164 & 514 & 517 \\
\hline
Total & 551,353 & 91,836 & 84,424\\
    \end{tabular}
    }
    \end{subtable} 
\end{table}
\subsubsection{Task Formulation: Forecasting TKGQA}
Forecasting TKGQA aims to test the forecasting ability of TKGQA models. It requires QA models to predict future facts based on the past TKG information. 
We formulate it as follows.
Given a TKG $\mathcal{G}$ and a natural language question $q$ generated based on a TKG fact whose valid timestamp is $t_q$, forecasting TKGQA aims to predict the answer to $q$. 
We label every question $q$ with $t_q$, and constrain QA models to only use the TKG facts $\{(s_i,r_i,o_i,t_i)|t_i<t_q\}$ before $t_q$ for answer inference. We propose three types of forecasting TKGQA questions, i.e., EPQs, YUQs, and FRQs. The answer to a EPQ is an entity $e \in \mathcal{E}$. The answer to a YUQ is either \textit{yes} or \textit{unknown}. We formulate FRQs as multiple choices and thus the answer to an FRQ corresponds to a choice $c$.
As a novel task, forecasting TKGQA requires models to have the ability of both natural language understanding (NLU) and future forecasting. Compared with it, the traditional TKG forecasting task does not require NLU and non-forecasting TKGQA does not consider future forecasting. Thus, previous methods for TKG forecasting\footnote{Relation set is provided in TKG forecasting and these methods explicitly learn relation representations. However, TKG relations are not annotated in forecasting TKGQA questions. Only question texts are provided and these methods have no way to process. Therefore, we do not consider them in experiments on our new task.}, e.g., RE-N\textsc{et} \cite{DBLP:conf/emnlp/JinQJR20}, and non-forecasting TKGQA, e.g., TempoQR \cite{DBLP:conf/aaai/MavromatisSIAHG22}, are not suitable for solving forecasting TKGQA.

\section{F\textsc{orecast}TKGQ\textsc{uestions}}
\subsection{Temporal Knowledge Base}
\begin{table}[htbp]
\caption{ICEWS21 TKG statistics. $N_\text{train}$, $N_\text{valid}$, $N_\text{test}$ denote the number of TKG facts in $\mathcal{G}_\text{train}$, $\mathcal{G}_\text{valid}$, $\mathcal{G}_\text{test}$, respectively. $|\mathcal{E}|$, $|\mathcal{R}|$, $|\mathcal{T}|$ denote ICEWS21's number of entities, relations, timestamps, respectively.}
\label{tab: icews21}
\small
    \centering
    \resizebox{0.55\columnwidth}{!}{
\begin{tabular}{l |c| c| c| c| c| c} 
\textbf{Dataset}&$N_\text{train}$&$N_\text{valid}$&$N_\text{test}$&$|\mathcal{E}|$&$|\mathcal{R}|$&$|\mathcal{T}|$
\\ 
\hline
ICEWS21  & 252,434 & 43,033 & 39,836 & 20,575 & 253 & 243  \\
\end{tabular}}
\end{table}
A subset from ICEWS \cite{DVN/28075_2015} is taken as the associated temporal KB for our proposed dataset.
We construct a TKG ICEWS21 based on the events taken from the official website of the ICEWS weekly event data\footnote{https://dataverse.harvard.edu/dataverse/icews} \cite{DVN/28075_2015}. ICEWS contains socio-political events in english. We take the events from Jan. 1, 2021, to Aug. 31, 2021, and extract TKG facts in the following way. For every ICEWS event, we generate a TKG fact $(s,r,o,t)$. We take the content of \textit{Event Date} as the timestamp $t$ of the TKG fact. We take the contents of \textit{Source Name} and \textit{Target Name} as the subject entity $s$ and the object entity $o$ of the TKG fact, respectively. We take the content of \textit{Event Text} as the relation type $r$ of the fact. We present the dataset statistics of ICEWS21 in Table \ref{tab: icews21}.
We split ICEWS21 into three parts $\mathcal{G}_{\text{train}} = \{(s,r,o,t)\in\mathcal{G}|t\in [ \,t_0,t_1) \,\}$, $\mathcal{G}_{\text{valid}} = \{(s,r,o,t)\in\mathcal{G}|t\in [ \,t_1,t_2) \,\}$, $\mathcal{G}_{\text{test}} = \{(s,r,o,t)\in\mathcal{G}|t\in [ \,t_2,t_3] \,\}$, where $t_0$, $t_1$, $t_2$, $t_3$ correspond to \textit{2021-01-01}, \textit{2021-07-01}, \textit{2021-08-01} and \textit{2021-08-31}, respectively. We generate training/validation/test questions based on $\mathcal{G}_{\text{train}}$/$\mathcal{G}_{\text{valid}}$/$\mathcal{G}_{\text{test}}$. We ensure that there exists no temporal overlap between every two of them, i.e., $\mathcal{G}_{\text{train}} \cap \mathcal{G}_{\text{valid}} = \emptyset$, $\mathcal{G}_{\text{train}} \cap \mathcal{G}_{\text{test}} = \emptyset$ and $\mathcal{G}_{\text{valid}} \cap \mathcal{G}_{\text{test}} = \emptyset$. In this way, we prevent QA models from observing any information from the evaluation sets during training.
\subsection{Question Categorization and Generation}
\label{sec: question generation}
We generate natural language questions based on the TKG facts in ICEWS21 and propose our QA dataset F\textsc{orecast}TKGQ\textsc{uestions}. 
Every relation type in ICEWS21 is coupled with a CAMEO code (specified in the \textit{CAMEO Code} column of the ICEWS weekly event data). In the official CAMEO codebook (can be found in ICEWS database), each CAMEO code is explained with examples and detailed descriptions.
We use the official CAMEO codebook provided in the ICEWS dataverse for aiding the generation of natural language relation templates. We create relation templates for 250 out of 253 relation types for question generation\footnote{The rest three relation types are not ideal for question generation (Appendix C.1).}. For example, we create a relation template \textit{engage in material cooperation with} for the relation type \textit{engage in material cooperation, not specified below}.
Questions in F\textsc{orecast}TKGQ\textsc{uestions} are categorized into three categories, i.e., EPQs (including 1-hop and 2 hop EPQs), YUQs, and FRQs. We summarize the numbers of different types of questions in Table \ref{tab: dataset statistics}. We use the relation templates to create natural language question templates for all types of questions (examples in Table \ref{tab: question template example}) which are used for question generation. All question templates are presented in our supplementary source code and explained in Appendix C.2. Same as previous KGQA datasets, e.g., C\textsc{ron}Q\textsc{uestions}, entity linking is considered as a separate problem and is not covered in our work. We assume complete entity and timestamp linking, and annotate the entities and timestamps in our questions. This applies to all three types of questions in our dataset. Distribution of question timestamps is specified in Appendix C.5.
\begin{table}[t]
\caption{Example question templates of all types. $s_q$ and $o_q$ are the annotated question entities. $t_q$ is the annotated question timestamp. For FRQ, $\{s_c\}$, $\{o_c\}$, $\{t_c\}$ are annotated choice entities and timestamp. We only write one choice in FRQ template for brevity. Better understand with details in Section \ref{sec: question generation}.}
\label{tab: question template example}
\small
    \centering
    \resizebox{\columnwidth}{!}{
\begin{tabular}{c| c} 
\textbf{Question Type}&\textbf{Example Template}
\\ 
\hline
1-Hop EPQ  &\textit{Who will $\{s_q\}$ engage in material cooperation with on $\{t_q\}$?}\\
\hline
2-Hop EPQ  &\textit{Who will threaten a country, while $\{s_q\}$ criticizes or denounces this country on $\{t_q\}$?} \\
\hline
YUQ & \textit{Will $\{s_q\}$ make a pessimistic comment about $\{o_q\}$ on $\{t_q\}$?}\\
\hline
\multirow{2}{*}{FRQ} & \textit{Why will $\{s_q\}$ appeal to $\{o_q\}$ to meet or negociate on $\{t_q\}$?} \\
& \textit{A: $\{s_c\}$ threaten $\{o_c\}$ on $\{t_c\}$; B:...}\\

\end{tabular}
}
\end{table}
\subsubsection{Entity Prediction Questions} 
We generate two groups of EPQs, i.e., 1-hop and 2-hop EPQs. 
Each 1-hop EPQ is generated from a single TKG fact, e.g., the natural language question \textit{Who will \underline{Sudan} host on \underline{2021-08-01}?} is based on (\textit{Sudan}, \textit{host}, \textit{Ramtane Lamamra}, \textit{2021-08-01}). Question templates are used during question generation. The underlined parts in the question denote the annotated entities and timestamps for KGQA. We consider all the facts concerning the 250 selected relations and transform them into 1-hop EPQs.
Each 2-hop EPQ is generated from two associated TKG facts in ICEWS21 where they contain common entities. 
An example is presented in Table \ref{tab: 2-hop EPQ example}.
The answer to a 2-hop EPQ (\textit{Israel}) corresponds to a 2-hop neighbor of its annotated entity (\textit{Iran}) at the question timestamp (\textit{2021-08-02}). We generate 2-hop questions by utilizing AnyBURL \cite{https://doi.org/10.48550/arxiv.2004.04412}, a rule-based KG reasoning model. We first split ICEWS21 into snapshots, where each snapshot $\mathcal{G}_{t_i} = \{(s,r,o,t)\in \mathcal{G}|t = t_i\}$ contains all the TKG facts happening at the same timestamp. Then we train AnyBURL on each snapshot for rule extraction. We collect the 2-hop rules with a confidence higher than 0.5 returned by AnyBURL, and manually check if two associated TKG facts in each rule potentially have a logical causation or can be used to interpret positive/negative entity relationships. After excluding the rules not meeting this requirement, we create question templates based on the remaining ones. We search for the groundings in ICEWS21 at every timestamp, where each grounding corresponds to a 2-hop EPQ. See our source code for the complete list of extracted 2-hop rules and see Appendix C.3 for more EPQ generation details.
\subsubsection{Yes-Unknown Questions}
Based on the idea of triple classification in KG reasoning\footnote{For a KG fact $(s,r,o)$, triple classification aims to predict whether this fact is valid or not.}, we introduce yes-no questions 
into KGQA. We then turn yes-no questions into yes-unknown questions because according to the Open World Assumption (OWA), the facts not observed in a given TKG are not necessarily wrong \cite{DBLP:conf/www/GalarragaTHS13}. We generalize triple classification to quadruple classification\footnote{Quadruple classification has never been studied in previous work. We define it as predicting whether a TKG fact $(s,r,o,t)$ is valid or unknown, under OWA.}, and then translate TKG facts into natural language questions. We take answering YUQs as solving quadruple classification. For every TKG fact concerning the selected 250 relations, we generate either a true or an unknown question based on it. For example, for the fact (\textit{Sudan}, \textit{host}, \textit{Ramtane Lamamra}, \textit{2021-08-01}), a true question is generated as \textit{Will \underline{Sudan} host \underline{Ramtane Lamamra} on \underline{2021-08-01}?} and we label \textit{yes} as its answer. An unknown question is generated by randomly perturbing one entity or the relation type in this fact, e.g., \textit{Will \underline{Germany} host \underline{Ramtane Lamamra} on \underline{2021-08-01}?}, and we label \textit{unknown} as its answer. We ensure that the perturbed fact does not exist in the original TKG. We use 25\% of total facts in ICEWS21 to generate true questions and the rest are used to generate unknown questions.
\begin{table}[t]
\caption{2-hop EPQ example. To avoid overlong text, we use symbols to represent relations and timestamps in TKG facts and 2-hop rules. $r_1 = $\textit{accuse}; $r_2 = $\textit{engage in diplomatic cooperation}; $t_1 = $\textit{2021-08-02}. $m, n$ are two entities that are 2-hop neighbors of each other at $t_1$. $X$ is their common 1-hop neighbor at $t_1$. The extracted rule describes the negative relationship between \textit{Iran} and \textit{Israel}.}
\label{tab: 2-hop EPQ example}
\small
    \centering
    \resizebox{\columnwidth}{!}{
\begin{tabular}{c| c| c| c} 
\textbf{Associated TKG Facts}&\textbf{2-Hop Rule}&\textbf{Generated 2-Hop Question}&\textbf{Answer}
\\ 
\hline
(\textit{United States}, $r_1$, \textit{Iran}, $t_1$)  &	$(X, r_1, m)$ & \textit{Who will a country engage in diplomatic cooperation with,} & \textit{Israel}  \\

(\textit{United States}, $r_2$, \textit{Israel}, $t_1$)& $=> (X, r_2, n)$ &\textit{while this country accuses \underline{Iran} on \underline{2021-08-02}?}& \\

\end{tabular}
}
\end{table}
\subsubsection{Fact Reasoning Questions}
\begin{figure*}[ht!]
    \centering
    \includegraphics[width=0.9\textwidth]{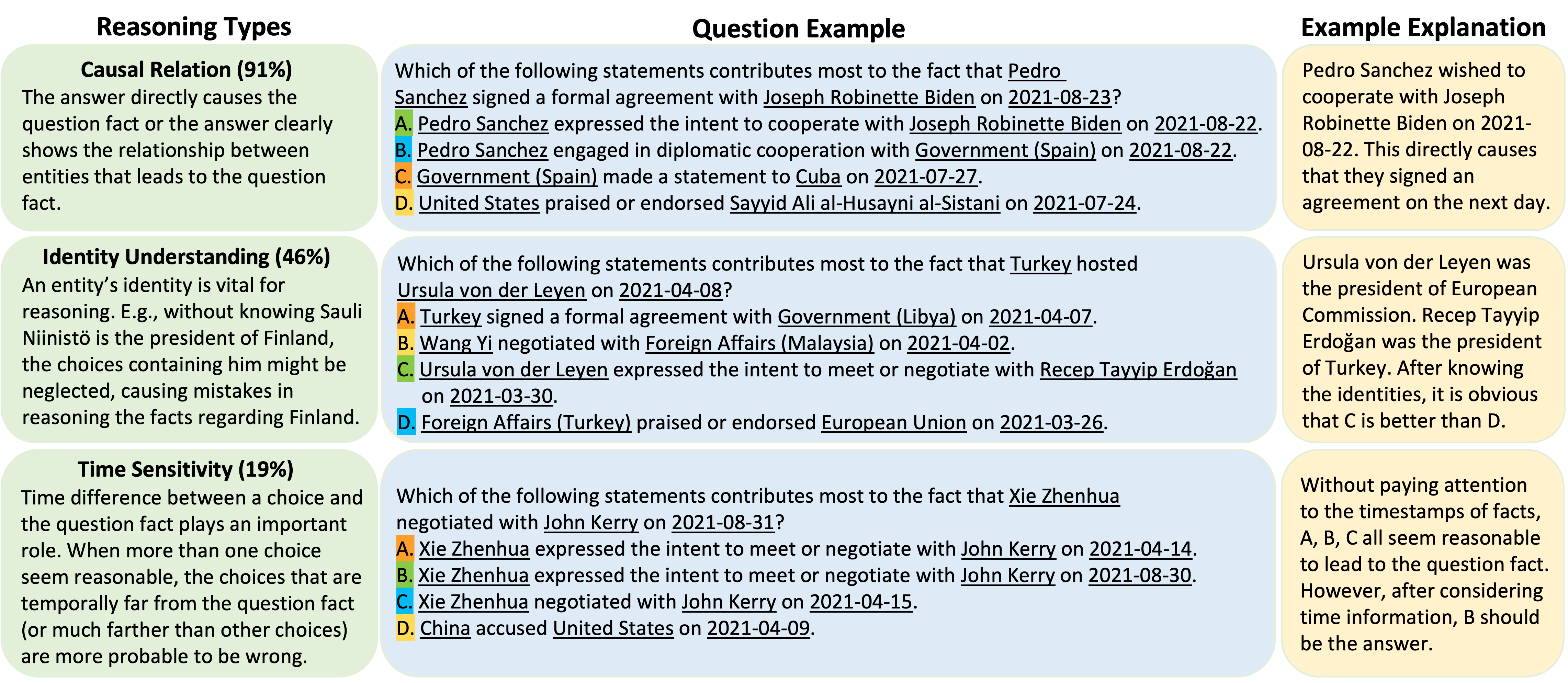}
    \caption{Required reasoning types and proportions (\%) in sampled FRQs, as well as FRQ examples. We sample 100 FRQs in each train/valid/test set. For choices, green for \textbf{Answer}, blue for \textbf{Hard Negative}, orange for \textbf{Median} and yellow for \textbf{Negative}. Multiple reasoning skills are required to answer each question, so the total proportion sum is not 100\%.}
    \label{fig: mc example}
\end{figure*}

The motivation for proposing FRQs is to study the difference between humans and machines in finding the supporting evidence for reasoning. We formulate FRQs in the form of multiple choices. Each question is coupled with four choices. Given a TKG fact from an FRQ, we ask the QA models to choose which fact in the choices is the most contributive to (the most relevant cause of) the fact mentioned in the question. We provide several examples in Fig. \ref{fig: mc example}.
We generate FRQs as follows. We first train a TKG forecasting model xERTE \cite{DBLP:conf/iclr/HanCMT21} on ICEWS21. Note that to predict a query $(s,r,?,t)$, xERTE samples its related prior TKG facts and assigns contribution scores to them. It provides explainability by assigning higher scores to the more related prior facts. We perform TKG forecasting and collect the queries where the ground-truth missing entities are ranked as top 1 by xERTE. For each collected query, we find its corresponding TKG fact 
and pick out four related prior facts found by xERTE. We take the prior facts with the highest, the lowest, and median contribution scores as \textbf{Answer}, \textbf{Negative}, and \textbf{Median}, respectively. Inspired by InferWiki \cite{DBLP:conf/acl/0002JLL0Z20}, we include a \textbf{Hard Negative} fact with the second highest contribution score, making it non-trivial for QA models to make the right decision. 
We generate each FRQ by turning the corresponding facts into a question and four choices (using templates), and manage to use xERTE to generate a large number of questions. However, since the answers to these questions are solely determined by xERTE, there exist numerous erroneous examples.
For example, the \textbf{Hard Negative} of lots of them are more suitable than their \textbf{Answer} to be the answers. 
We ask five graduate students (major in computer science) to manually check all these questions and annotate them as reasonable or unreasonable according to their own knowledge or through search engines. If the majority annotate a question as unreasonable, we filter it out. See Appendix C.4 for more details of FRQ generation and annotation, including the annotation instruction and interface.

To better study the reasoning skills required to answer FRQs, we randomly sample 300 FRQs and manually annotate them with reasoning types. The required reasoning skills and their proportions are shown in Fig. \ref{fig: mc example}.
\section{F\textsc{orecast}TKGQA}
\label{sec: model}
\begin{figure*}[t]
    \centering
    \includegraphics[width=0.9\textwidth]{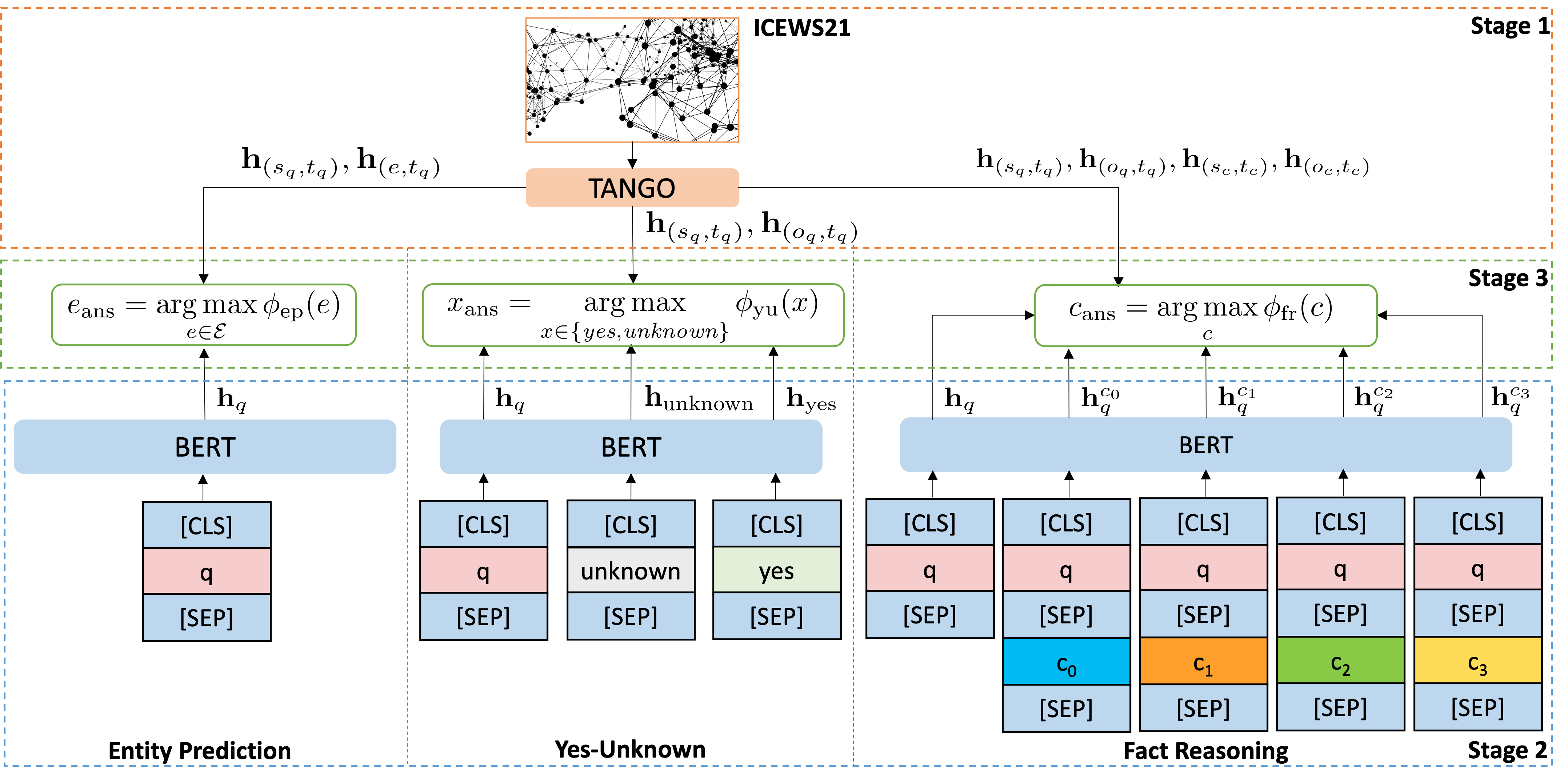}
    \caption{Model structure of F\textsc{orecast}TKGQA.}
    \label{fig: model structure}
\end{figure*}

F\textsc{orecast}TKGQA employs a TKG forecasting model TANGO \cite{DBLP:conf/emnlp/HanDMGT21} and a pre-trained LM BERT \cite{DBLP:conf/naacl/DevlinCLT19} for solving forecasting questions. We illustrate its model structure in Fig. \ref{fig: model structure} with three stages. In Stage 1, a TKG forecasting model TANGO \cite{DBLP:conf/emnlp/HanDMGT21} is used to generate the time-aware representation for each entity at each timestamp. In Stage 2, a pre-trained LM (e.g., BERT) is used to encode questions (and choices) into question (choice) representations. Finally, in Stage 3, answers are predicted according to the scores computed using the representations from Stage 1 and 2.
\subsection{TKG Forecasting Model}
We train TANGO on ICEWS21 with the TKG forecasting task. We use ComplEx \cite{DBLP:conf/icml/TrouillonWRGB16} as its scoring function. We learn the entity and relation representations in the complex space $\mathbb{C}^d$, where $d$ is the dimension of complex vectors. The training set corresponds to all the TKG facts in $\mathcal{G}_{\text{train}}$, and we evaluate the trained model on $\mathcal{G}_{\text{valid}}$ and $\mathcal{G}_{\text{test}}$. After training, we perform a one time inference on $\mathcal{G}_{\text{valid}}$ and $\mathcal{G}_{\text{test}}$. 
Following the default setting of TANGO, to compute entity and relation representations at every timestamp $t$, we recurrently input all the TKG facts from $t-4$ to $t-1$, i.e., snapshots from $\mathcal{G}_{t-4}$ to $\mathcal{G}_{t-1}$, into TANGO and take the output representations.
Note that it infers representations based on the prior facts, thus not violating our forecasting setting. We compute the entity and relation representations at every timestamp in ICEWS21 and keep them for aiding the QA systems in Stage 1 (Fig. \ref{fig: model structure}). See Appendix B.1 for more details of TANGO training and inference.
To leverage the complex representations computed by TANGO with ComplEx, we map the output of BERT to $\mathbb{C}^d$. 
For each natural language input, we take the output representation of the [CLS] token computed by BERT and project it to a 2$d$ real space to form a 2$d$ real valued vector. We take the first and second half of it as the real and imaginary part of a $d$-dimensional complex vector, respectively.
All the representations output by BERT have already been mapped to $\mathbb{C}^d$ without further notice.
\subsection{QA Model}
\subsubsection{Entity Prediction}
For every EPQ $q$, we compute an entity score for every entity $e \in \mathcal{E}$. The entity with the highest score is predicted as the answer $e_\text{ans}$. To compute the score for $e$, we first input $q$ into BERT and map its output to $\mathbb{C}^d$ to get the question representation $\mathbf{h}_q$. Inspired by ComplEx, we then define $e$'s entity score as
\begin{equation}
    \label{eq: es}
    \phi_{\text{ep}}(e) = \text{Re}\left(<\mathbf{h}'_{(s_q,t_q)},\mathbf{h}_q, \Bar{\mathbf{h}}'_{(e,t_q)}>\right).
\end{equation}
$\mathbf{h}'_{(s_q,t_q)} = f_\text{ep}\left(\mathbf{h}_{(s_q,t_q)}\right)$, $\mathbf{h}'_{(e,t_q)} = f_\text{ep}\left(\mathbf{h}_{(e,t_q)}\right)$, where $f_\text{ep}$ denotes a neural network aligning TKG representations to EPQs.  $\mathbf{h}_{(s_q,t_q)}$ and $\mathbf{h}_{(e,t_q)}$ denote the TANGO representations of the annotated entity $s_q$ and the entity $e$ at the question timestamp $t_q$, respectively. \text{Re} means taking the real part of a complex vector and $\Bar{\mathbf{h}}'_{(e,t_q)}$ means the complex conjugate of $\mathbf{h}'_{(e,t_q)}$.

\subsubsection{Yes-Unknown Judgment}
For a YUQ, we compute a score for each candidate answer $x \in \{\textit{yes}, \textit{unknown}\}$.
We first encode each $x$ into a $d$-dimensional complex representation $\mathbf{h}_x$ with BERT. Inspired by TComplEx \cite{DBLP:conf/iclr/LacroixOU20}, we then compute scores as
\begin{equation}
\label{eq: yns}
    \phi_{\text{yu}}(x) = \text{Re}\left(<\mathbf{h}'_{(s_q,t_q)},\mathbf{h}_q, \Bar{\mathbf{h}}'_{(o_q,t_q)}, \mathbf{h}_x>\right).
\end{equation}
$\mathbf{h}'_{(s_q,t_q)} = f_\text{yu}\left(\mathbf{h}_{(s_q,t_q)}\right)$,$\mathbf{h}'_{(o_q,t_q)} = f_\text{yu}\left(\mathbf{h}_{(o_q,t_q)}\right)$, where $f_\text{yu}$ denotes a neural network aligning TKG representations to YUQs. $\mathbf{h}_{(s_q,t_q)}$ and $\mathbf{h}_{(o_q,t_q)}$ denote the TANGO representations of the annotated subject entity $s_q$ and object entity $o_q$ at $t_q$, respectively. $\mathbf{h}_q$ is the BERT encoded question representation. 
We take the candidate answer with the higher score as the predicted answer $x_\text{ans}$.

\subsubsection{Fact Reasoning}
We compute a choice score for every choice $c$ in an FRQ by using the following scoring function:
\begin{equation}
\label{eq: cs}
    \phi_{\text{fr}}(c) = \text{Re}\left(<\mathbf{h}'_{(s_c,t_c)},\mathbf{h}^{c}_q, \Bar{\mathbf{h}}'_{(o_c,t_c)}, \mathbf{h}'_q>\right),
\end{equation}
$\mathbf{h}^{c}_q$ is the output of BERT mapped to $\mathbb{C}^d$ given the concatenation of $q$ and $c$. $\mathbf{h}'_{(s_c,t_c)} = f_\text{fr}\left(\mathbf{h}_{(s_c,t_c)}\right)$ and $\mathbf{h}'_{(o_c,t_c)} = f_\text{fr}\left(\mathbf{h}_{(o_c,t_c)}\right)$. $f_\text{fr}$ is a projection network and $\mathbf{h}_{(s_c,t_c)}$, $\mathbf{h}_{(o_c,t_c)}$ denote the TANGO representations of the entities annotated in $c$. 
$\mathbf{h}'_q = f\left(f_\text{fr}\left(\mathbf{h}_{(s_q,t_q)}\right)\|\mathbf{h}^{c}_{q}\|f_\text{fr}\left(\mathbf{h}_{(o_q,t_q)}\right)\right)$, where $f$ serves as a projection and $\|$ denotes concatenation. $\mathbf{h}_{(s_q,t_q)}$ and $\mathbf{h}_{(o_q,t_q)}$ denote the TANGO representations of the entities annotated in the question $q$. 
We take the choice with the highest choice score as our predicted answer $c_\text{ans}$. We give a more detailed description of Equation \ref{eq: es}, \ref{eq: yns} and \ref{eq: cs} in Appendix A.

\subsubsection{Parameter Learning}
We use cross-entropy loss to train F\textsc{orecast}TKGQA on each type of questions separately. The loss functions of EPQs, FRQs and YUQs are given by $\mathcal{L}_{\text{ep}} = -  \sum_{q \in \mathcal{Q}^{\text{ep}}} \log \left( \frac{\phi_{\text{ep}}(e_\text{ans})}{\sum_{e \in \mathcal{E}} \phi_{\text{ep}}(e)} \right)$, $\mathcal{L}_{\text{fr}} = - \sum_{q \in \mathcal{Q}^{\text{fr}}} \log \left( \frac{\phi_{\text{fr}}(c_\text{ans})}{\sum_{c} \phi_{\text{fr}}(c)} \right)$ and $\mathcal{L}_{\text{yu}} = -  \sum_{q \in \mathcal{Q}^{\text{yu}}} \log \left( \frac{\phi_{\text{yu}}(x_\text{ans})}{\sum_{x \in \{\textit{yes}, \textit{unknown}\}} \phi_{\text{yu}}(x)} \right)$, respectively.
$\mathcal{Q}^\text{ep}$/$\mathcal{Q}^\text{yu}$/$\mathcal{Q}^\text{fr}$ denotes all EPQs/YUQs/FRQs and $e_\text{ans}$/$x_\text{ans}$/$c_\text{ans}$ is the answer to question $q$. 

\section{Experiments}
We answer several research questions (RQs) with experiments\footnote{Implementation details and further analysis of F\textsc{orecast}TKGQA in Appendix B.3 and G.}.
\textbf{RQ1} (Section \ref{sec: main results}, \ref{sec: Performance over FRQs with Different Reasoning Types}): Can a TKG forecasting model better support forecasting TKGQA than a TKGC model? \textbf{RQ2} (Section \ref{sec: main results}, \ref{sec: Performance over FRQs with Different Reasoning Types}): Does F\textsc{orecast}TKGQA perform well in forecasting TKGQA? \textbf{RQ3} (Section \ref{sec: human}, \ref{sec: answerability}): Are the questions in our dataset answerable? \textbf{RQ4} (Section \ref{sec:appendix data efficiency}): Is the proposed dataset efficient? \textbf{RQ5} (Section \ref{sec: challenges}): What are the challenges of forecasting TKGQA?
\subsection{Experimental Setting}
\subsubsection{Evaluation Metrics}
We use mean reciprocal rank (MRR) and Hits@k as the evaluation metrics of the EPQs. For each EPQ, we compute the rank of the ground-truth answer entity among all the TKG entities. Test MRR is then computed as $\frac{1}{|\mathcal{Q}^\text{ep}_\text{test}|}\sum_{q \in \mathcal{Q}^\text{ep}_\text{test}} \frac{1}{\text{rank}_q}$, where $\mathcal{Q}^\text{ep}_\text{test}$ denotes all EPQs in the test set and $\text{rank}_q$ is the rank of the ground-truth answer entity of question $q$. Hits@k is the proportion of the answered questions where the ground-truth answer entity is ranked as top k. For YUQs and FRQs, we employ accuracy for evaluation. Accuracy is the proportion of the correctly answered questions out of all questions.
\subsubsection{Baseline Methods}
We consider two pre-trained LMs, BERT \cite{DBLP:conf/naacl/DevlinCLT19} and RoBERTa \cite{https://doi.org/10.48550/arxiv.1907.11692} as baselines. For EPQs and YUQs, we add a prediction head on top of the question representations computed by LMs, and use softmax function to compute answer probabilities. For every FRQ, we input into each LM the concatenation of the question with each choice, and follow the same prediction structure. Besides, we derive two model variants for each LM by introducing TKG representations. We train TComplEx on ICEWS21. For every EPQ and YUQ, we concatenate the question representation with the TComplEx representations of the entities and timestamps annotated in the question, and then perform prediction with a prediction head and softmax. For FRQs, we further include TComplEx representations into choices in the same way. We call this type of variant BERT\_int and RoBERTa\_int since TComplEx is a TKGC (TKG interpolation) method. Similarly, we also introduce TANGO representations into LMs and derive BERT\_ext and RoBERTa\_ext, where TANGO serves as a TKG extrapolation backend. Detailed model derivations are presented in Appendix B.2.
We also consider one KGQA method EmbedKGQA \cite{saxena-etal-2020-improving}, and two TKGQA methods, i.e., C\textsc{ron}KGQA \cite{DBLP:conf/acl/SaxenaCT20} and TempoQR \cite{DBLP:conf/aaai/MavromatisSIAHG22} as baselines. We run EmbedKGQA on top of the KG representations trained with ComplEx 
on ICEWS21, and run TKGQA baselines on top of the TKG representations trained with TComplEx. 
\begin{table}[t]
    \caption{Experimental results over F\textsc{orecast}TKGQ\textsc{uestions}. The best results are marked in bold.}
    \label{tab: main results}
    \centering
    \begin{subtable}{.5\linewidth}
      \centering
        \caption{EPQs. Overall results in Appendix D.} \label{tab: main results ep}
            \resizebox{\columnwidth}{!}{
    \begin{tabular}{@{}lcc cc cc@{}}
\toprule
         & \multicolumn{2}{c}{\textbf{MRR}} & \multicolumn{2}{c}{\textbf{Hits@1}} & \multicolumn{2}{c}{\textbf{Hits@10}}\\
\cmidrule(lr){2-3} \cmidrule(lr){4-5} \cmidrule(lr){6-7}
\textbf{Model}  & 1-Hop & 2-Hop  & 1-Hop & 2-Hop  & 1-Hop & 2-Hop\\

\midrule
        RoBERTa  & 0.166 & 0.149 
         & 0.104 & 0.085 
         & 0.288 & 0.268 
         \\
        BERT  & 0.279 & 0.182 
         & 0.192 & 0.106 
         & 0.451 & 0.342 
         \\
\midrule
        EmbedKGQA  & 0.317 & 0.185 
         & 0.228 & 0.112 
         & 0.489 & 0.333 
        \\
\midrule 
        RoBERTa\_int  & 0.283 & 0.157 
         & 0.190 & 0.094 
         & 0.467 & 0.290 
         \\
        BERT\_int  & 0.314 & 0.183 
         & 0.223 & 0.107 
         & 0.490 & 0.344
        \\
        C\textsc{ron}KGQA  & 0.131 & 0.090 
         & 0.081 & 0.042
         & 0.231 & 0.187 
         \\
        TempoQR  & 0.145 & 0.107 
         & 0.094 & 0.061 
         & 0.243 & 0.199 
         \\
\midrule
        RoBERTa\_ext  & 0.306 & 0.180
         & 0.216 & 0.108
         & 0.497 & 0.323 
         \\
        BERT\_ext  & 0.331 & 0.208 
         & 0.239 & 0.128 
         & 0.508 & 0.369
        \\    
\midrule
        F\textsc{orecast}TKGQA 
         & \textbf{0.339}&\textbf{0.216} 
         & \textbf{0.248}&\textbf{0.129} 
         & \textbf{0.517}&\textbf{0.386}
        \\
\bottomrule
    \end{tabular}
    }
    \end{subtable}%
    \begin{subtable}{.355\linewidth}
      \centering
        \caption{YUQs and FRQs.} \label{tab: main results yu fr}
            \resizebox{\columnwidth}{!}{
    \begin{tabular}{@{}lcc@{}}
\toprule
     & \multicolumn{2}{c}{\textbf{Accuracy}}\\
\cmidrule(lr){2-3}
\textbf{Model} & YUQ & FRQ\\

\midrule
        RoBERTa & 0.721 & 0.645
         \\
        BERT & 0.813 & 0.634
         \\
\midrule 
        RoBERTa\_int & 0.768 & 0.693
         \\
        BERT\_int & 0.829 & 0.682
        \\
\midrule
        RoBERTa\_ext & 0.798 & 0.707 
         \\
        BERT\_ext & 0.837 & 0.746
         \\
\midrule
        F\textsc{orecast}TKGQA 
        & \textbf{0.870} & \textbf{0.769}
        \\
        Human Performance (a) & - & 0.936
        \\
        Human Performance (b) & - & 0.954
        \\
\bottomrule
    \end{tabular}
    }
    \end{subtable} 
\end{table}
\subsection{Main Results}
\label{sec: main results}
We report the experimental results in Table \ref{tab: main results}. In Table \ref{tab: main results ep}, we show that our entity prediction model outperforms all baseline methods. We observe that EmbedKGQA achieves a better performance than BERT and RoBERTa, showing that employing KG representations helps TKGQA. Besides, LM variants outperform their original LMs, indicating that TKG representations help LMs perform better in TKGQA. Further, BERT\_ext shows stronger performance than BERT\_int (this also applies to RoBERTa\_int and RoBERTa\_ext), which proves that TKG forecasting models provide greater help than TKGC models in forecasting TKGQA. C\textsc{ron}KGQA and TempoQR employ TComplEx representations as supporting information and perform poorly, implying that employing TKG representations provided by TKGC methods may include noisy information in forecasting TKGQA. 
F\textsc{orecast}TKGQA injects TANGO representations into a scoring module, showing its great effectiveness on EPQs. For YUQs and FRQs, F\textsc{orecast}TKGQA also achieves the best performance. Table \ref{tab: main results yu fr} shows that it is helpful to include TKG representations for answering YUQs and FRQs and our scoring functions 
are effective.
\subsection{Human vs. Machine on FRQs}
\label{sec: human}
To study the difference between humans and models in fact reasoning, we further benchmark human performance on FRQs with a survey (See Appendix E for details). We ask five graduate students to answer 100 questions randomly sampled from the test set. We consider two settings: (a) Humans answer FRQs with their own knowledge and inference ability. \textbf{Search engines are not allowed}; (b) Humans can turn to search engines and use the web information published \textbf{before the question timestamp} for aiding QA. Table \ref{tab: main results} shows that humans achieve much stronger performance than all QA models (even in setting (a)). This calls for a great effort to build better fact reasoning TKGQA models.
\subsection{Performance over FRQs with Different Reasoning Types}
\label{sec: Performance over FRQs with Different Reasoning Types}
\begin{figure}[ht!]
\centering

\subfloat[Case 1.]{%
  \includegraphics[width=0.5\columnwidth]{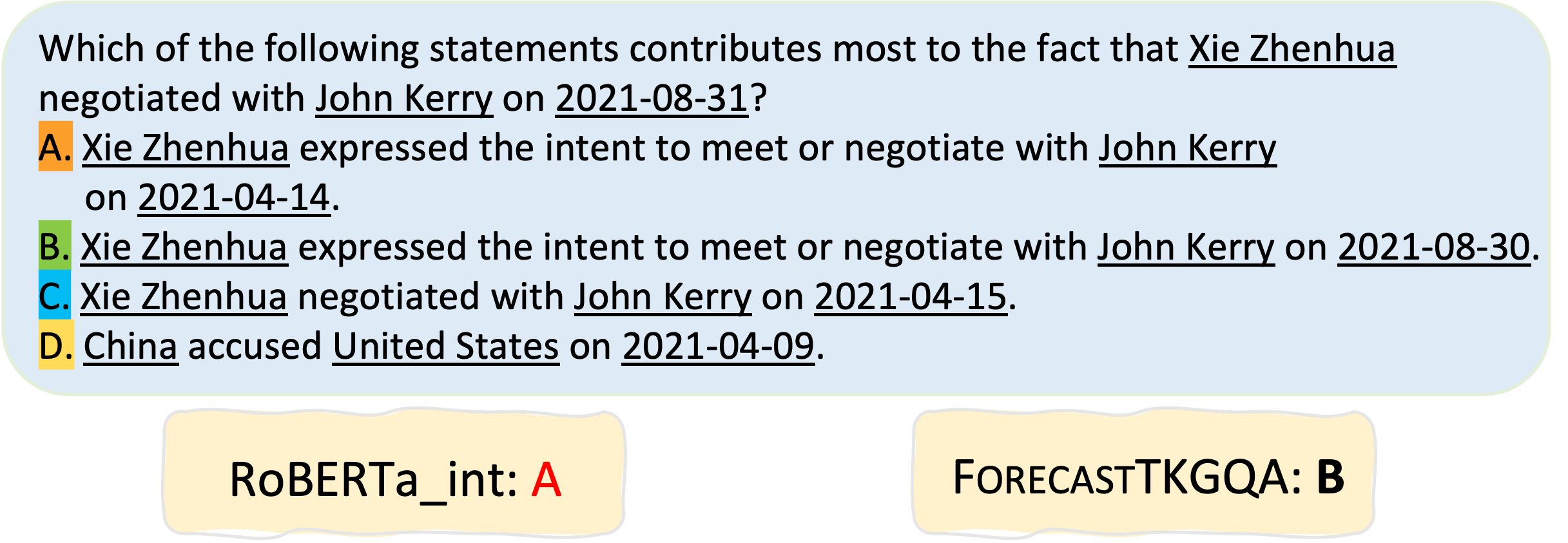}%
  \label{fig:case1}%
}\hfill
\subfloat[Case 2.]{%
  \includegraphics[width=0.5\columnwidth]{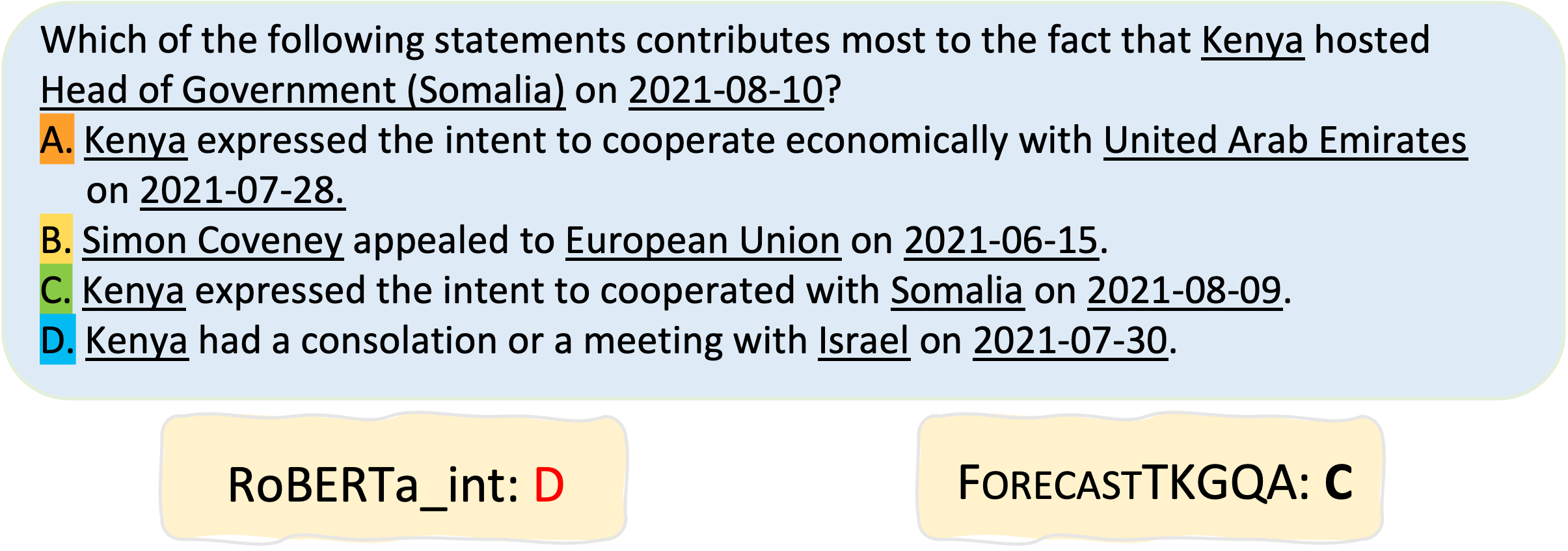}%
  \label{fig:case2}%
}

\caption{Case Studies on FRQs. We mark green for \textbf{Answer}, blue for \textbf{Hard Negative}, orange for \textbf{Median} and yellow for \textbf{Negative}.}\label{fig: case study frq}
\end{figure}

Considering the reasoning types listed in Fig. \ref{fig: mc example}, we compare RoBERTa\_int with F\textsc{orecast}TKGQA on the 100 sampled test questions that are annotated with reasoning types, to justify performance gain brought by TKG forecasting model on FRQs. 
Experimental results in Table \ref{tab: reason type result} imply that employing TKG forecasting model helps QA models better deal with any reasoning type on FRQs. We use two cases in Fig. \ref{fig: case study frq} to provide insights of performance gain.
\begin{table}[t]
\caption{Performance comparison across FRQs with different reasoning types.}\label{tab: reason type result}
\small
    \centering
    \resizebox{0.7\columnwidth}{!}{
    \begin{tabular}{@{}lccc@{}}
\toprule
     & \multicolumn{3}{c}{\textbf{Accuracy}}\\
\cmidrule(lr){2-4}
\textbf{Model} & Causal Relation & Identity Understanding & Time Sensitivity\\

\midrule 
        RoBERTa\_int & 0.670 & 0.529 & 0.444
         \\
\midrule
        F\textsc{orecast}TKGQA 
        & \textbf{0.787} & \textbf{0.735}  & \textbf{0.611}\\

\bottomrule
    \end{tabular}
    }
\end{table}
\paragraph{Case 1.} Two reasoning skills, i.e., Causal Relation and Time Sensitivity (shown in Fig. \ref{fig: mc example}), are required to correctly answer the question in Case 1. Without considering the timestamps of choices, A, B, C all seem at least somehow reasonable. However, after considering choice timestamps, B should be the most contributive reason for the question fact. First, the timestamp of B (\textit{2021-08-30}) is much closer to the question timestamp (\textit{2021-08-31}). Moreover, the fact in choice B directly causes the question fact. RoBERTa\_int manages to capture the causation, but fail to correctly deal with time sensitivity, while F\textsc{orecast}TKGQA achieves better reasoning on both reasoning types.
\paragraph{Case 2.} Two reasoning skills, i.e., Causal Relation and Identity Understanding (shown in Fig. \ref{fig: mc example}), are required to correctly answer the question in Case 2. 
\textit{Head of Government (Somalia)} and \textit{Somalia} are two different entities in TKG, however, both entities are about Somalia. By understanding this, we are able to choose the correct answer.
F\textsc{orecast}TKGQA manages to understand the identity of \textit{Head of Government (Somalia)}, match it with \textit{Somalia} and find the cause of the question fact. RoBERTa\_int makes a mistake because as a model equipped with TComplEx, it has no well-trained timestamp representations of the question and choice timestamps, which would introduce noise in decision making.

\subsection{Answerability of F\textsc{orecast}TKGQ\textsc{uestions}}
\label{sec: answerability}
To validate the answerability of the questions in F\textsc{orecast}TKGQ\textsc{uestions}. We train TComplEx and TANGO over the whole ICEWS21, i.e., $\mathcal{G}_{\text{train}}\cup\mathcal{G}_{\text{valid}}\cup\mathcal{G}_{\text{test}}$, and use them to support QA. Note that this violates the forecasting setting of forecasting TKGQA, and thus we call the TKG models trained in this way as cheating TComplEx (CTComplEx) and cheating TANGO (CTANGO).
Answering EPQs with cheating TKG models is same as non-forecasting TKGQA. We couple TempoQR with CTComplEx and see a huge performance increase (Table \ref{tab: answerability ep}). Besides, inspired by \cite{DBLP:conf/emnlp/JiKHWZH20}, we develop a new TKGQA model Multi-Hop Scorer\footnote{See Appendix F for detailed model explanation and model structure illustration.} (MHS) for EPQs. Starting from the annotated entity $s_q$ of an EPQ, MHS updates the scores of outer entities for $n$-hops ($n=2$ in our experiments) until all $s_q$'s $n$-hop neighbors on the snapshot $\mathcal{G}_{t_q}$ are visited. Initially, MHS assigns a score of 1 to $s_q$ and 0 to any other unvisited entity. 
For each unvisited entity $e$, it then computes $e$'s score as: 
$\phi_{\text{ep}}(e) = \frac{1}{|\mathcal{N}_e(t_q)|}\sum_{(e',r)\in\mathcal{N}_e(t_q)}(\gamma \cdot \phi_{\text{ep}}(e') + \psi(e',r,e,t_q))$, where $\mathcal{N}_e(t_q) = \{(e', r)|(e',r,e,t_q)\in \mathcal{G}_{t_q}\}$ is $e$'s 1-hop neighborhood on $\mathcal{G}_{t_q}$ and $\gamma$ is a discount factor. We couple MHS with CTComplEx and CTANGO, and define $\psi(e',r,e,t_q)$ separately. For MHS + CTComplEx, 
$\psi(e',r,e,t_q) = f_2(f_1(\mathbf{h}_{e'}\|\mathbf{h}_r\|\mathbf{h}_{e}\|\mathbf{h}_{t_q}\|\mathbf{h}_q))$.
$f_1$ and $f_2$ are two neural networks.
$\mathbf{h}_{e}, \mathbf{h}_{e'}, \mathbf{h}_{r}, \mathbf{h}_{t_q}$ are the CTComplEx representations of entities $e$, $e'$, relation $r$ and timestamp $t_q$, respectively. For MHS + CTANGO, we take the idea of F\textsc{orecast}TKGQA: $\psi(e',r,e,t_q) = \text{Re}\left(<\mathbf{h}_{(e',t_q)},\mathbf{h}_r, \Bar{\mathbf{h}}_{(e,t_q)}, \mathbf{h}_q>\right)$. 
$\mathbf{h}_{(e,t_q)}$, $\mathbf{h}_{(e',t_q)}$, $\mathbf{h}_r$ are the CTANGO representations of entities $e$, $e'$ at $t_q$, and relation $r$, respectively.
$\mathbf{h}_q$ is BERT encoded question representation. We find that MHS achieves superior performance (even on 2-hop EPQs). This is because MHS not only uses cheating TKG models, but also considers ground-truth multi-hop structural information of TKGs at $t_q$ (which is unavailable in the forecasting setting).
For YUQs and FRQs, Table \ref{tab: answerability yuq frq} shows that cheating TKG models help improve performance, especially on FRQs. 
These results imply that given the ground-truth TKG information at question timestamps, our forecasting TKGQA questions are answerable.
\begin{table}[t]
    \caption{Answerability study. Models with $\alpha$ means using CTComplEx and $\beta$ means using CTANGO. $\uparrow$ denotes relative improvement ($\%$) from the results in Table \ref{tab: main results}. Acc means Accuracy.
    }\label{tab: answerability}
    \begin{subtable}{.59\linewidth}
      \centering
        \caption{EPQs.}\label{tab: answerability ep}
\resizebox{\columnwidth}{!}{
    \begin{tabular}{@{}lcccc cccc@{}}
\toprule
         & \multicolumn{4}{c}{\textbf{MRR}} & \multicolumn{4}{c}{\textbf{Hits@10}}\\
\cmidrule(lr){2-5}\cmidrule(lr){6-9}
\textbf{Model}  & 1-Hop & $\uparrow$ & 2-Hop & $\uparrow$ & 1-Hop & $\uparrow$ & 2-Hop & $\uparrow$\\
\midrule 
        TempoQR$^\alpha$ & 0.713 & 391.7 & 0.233 & 117.8
         & 0.883 & 263.4 & 0.419 & 110.6
         \\
        MHS$^\alpha$
         & 0.868 & - & 0.647 & -
         & 0.992 & - & 0.904 & -
        \\
        MHS$^\beta$
         & 0.771 & - & 0.556 & -
         & 0.961 & - & 0.828 & -
        \\

\bottomrule
    \end{tabular}
    }
    \end{subtable}%
    \begin{subtable}{.41\linewidth}
      \centering
        \caption{YUQs and FRQs.}\label{tab: answerability yuq frq}
                \resizebox{\columnwidth}{!}{
    \begin{tabular}{@{}lcccc@{}}
\toprule
         & \multicolumn{2}{c}{\textbf{YUQ}} & \multicolumn{2}{c}{\textbf{FRQ}}\\
\cmidrule(lr){2-3}\cmidrule(lr){4-5}
\textbf{Model}  & Acc & $\uparrow$ & Acc & $\uparrow$ \\
\midrule 
        BERT\_int$^\alpha$ & 0.855 & 19.6
        & 0.816 & 14.4
         \\

        BERT\_ext$^\beta$ & 0.873 & 4.3
        & 0.836 & 12.1
        \\
        F\textsc{orecast}TKGQA$^\beta$ & 0.925 & 6.3
        & 0.821 & 6.8
        \\

\bottomrule
    \end{tabular}
    }
    \end{subtable} 
\end{table}
\subsection{Challenges of Forecasting TKGQA over F\textsc{orecast}TKGQ\textsc{uestions}}
\label{sec: challenges}
From the experiments discussed in Section \ref{sec: human} and \ref{sec: answerability}, we summarize the challenges of forecasting TKGQA: (1) Inferring the ground-truth TKG information $\mathcal{G}_{t_q}$ at the question timestamp $t_q$ accurately; (2) Effectively performing multi-hop reasoning for forecasting TKGQA; 
(3) Developing TKGQA models for better fact reasoning.
In Section \ref{sec: answerability}, we have trained cheating TKG models and used them to support QA. We show in Table \ref{tab: answerability} that QA models substantially improve their performance on forecasting TKGQA with cheating TKG models. This implies that accurately inferring the ground-truth TKG information at $t_q$ is crucial in our task and how to optimally achieve it remains a challenge. We also observe that MHS with cheating TKG models achieves much better results on EPQs (especially on 2-hop). MHS utilizes multi-hop information of the ground-truth TKG at $t_q$ ($\mathcal{G}_{t_q}$) for better QA.
In forecasting TKGQA, by only knowing the TKG facts before $t_q$ and not observing $\mathcal{G}_{t_q}$, it is impossible for MHS to directly utilize the ground-truth multi-hop information at $t_q$. This implies that how to effectively infer and exploit multi-hop information for QA in the forecasting scenario remains a challenge. Moreover, as discussed in Section \ref{sec: human}, current TKGQA models still trail humans with great margin on FRQs. It is challenging to design novel forecasting TKGQA models for better fact reasoning.
\subsection{Study of Data Efficiency}
\label{sec:appendix data efficiency}
\begin{figure}[ht!]
\centering
\subfloat[Data efficiency on EPQs.]{%
  \includegraphics[width=0.45\columnwidth]{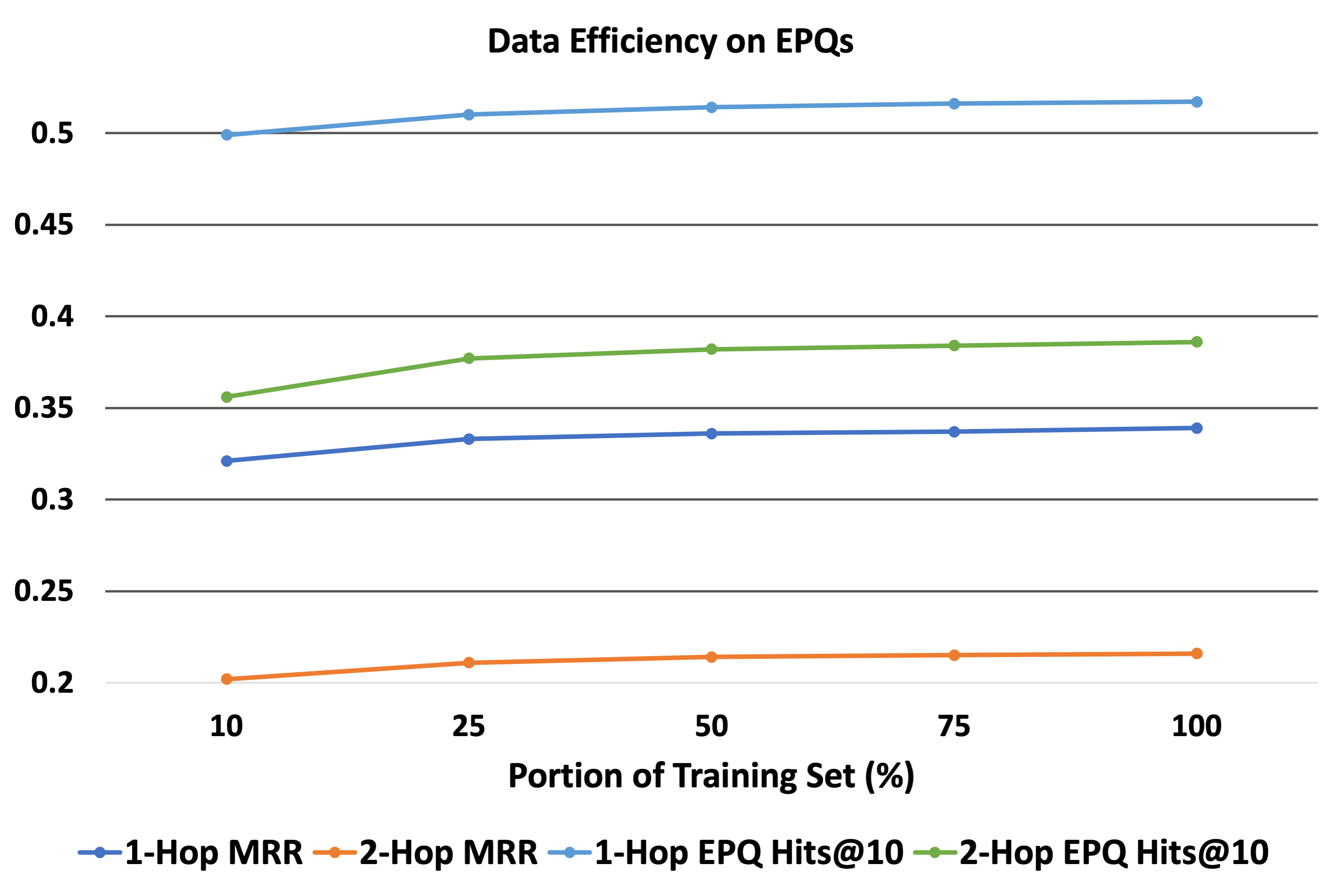}%
  \label{fig: data_eff_epq}%
}\hfill
\subfloat[Data efficiency on YUQs, FRQs.]{%
  \includegraphics[width=0.45\columnwidth]{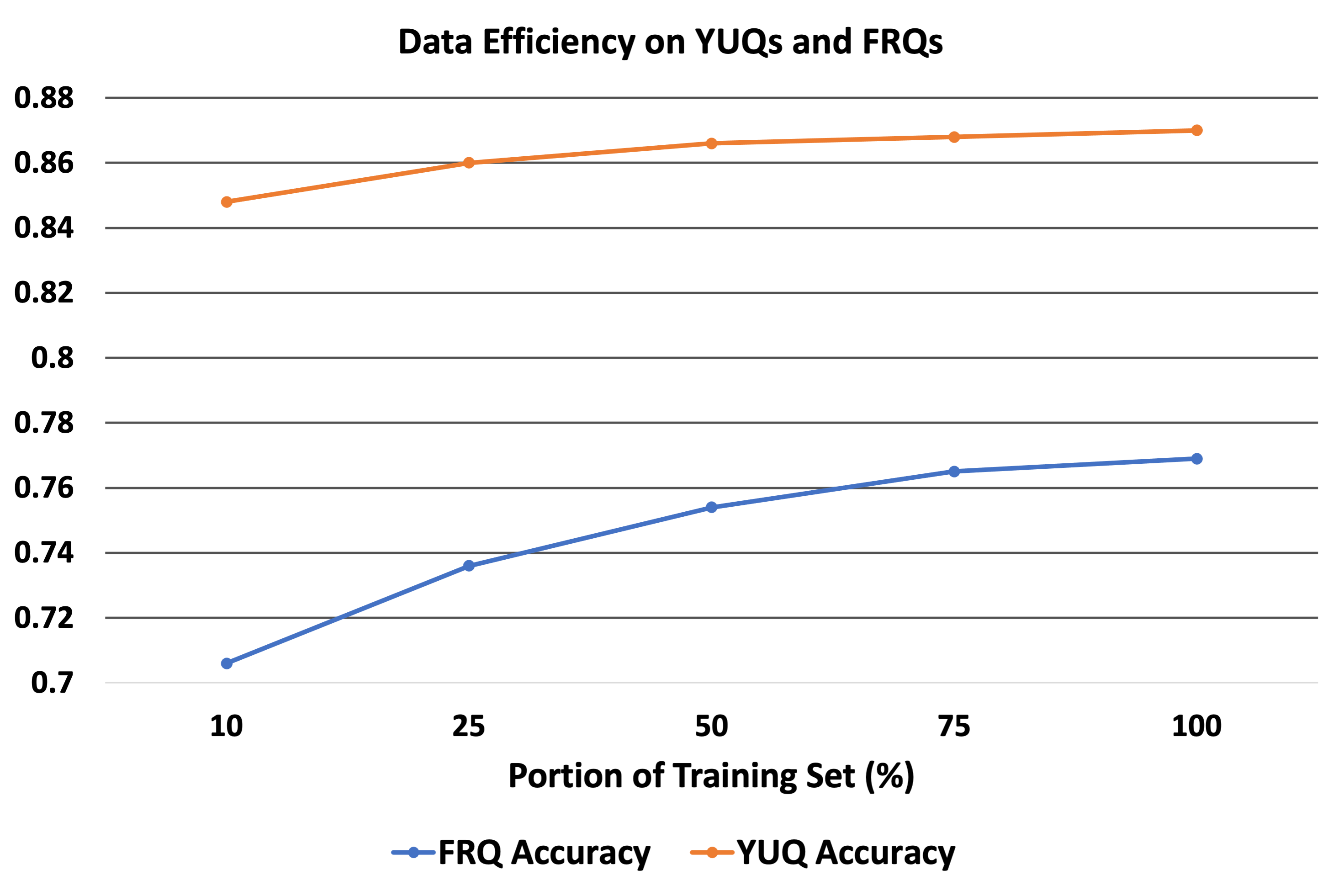}%
  \label{fig: data_eff_yuq_frq}%
}
\caption{Data efficiency analysis.}\label{fig: data_eff}
\end{figure}

We want to know how the models will be affected with less/more training data. For each type of questions, we modify the size of its training set. We train F\textsc{orecast}TKGQA on the modified training sets and evaluate our model on the original test sets. We randomly sample 10\%, 25\%, 50\%, and 75\% of the training examples to form new training sets. Fig. \ref{fig: data_eff} shows that for every type of question, the performance of F\textsc{orecast}TKGQA steadily improves as the size of the training sets increase. This proves that our proposed dataset is efficient and useful for training forecasting TKGQA models.

\section{Justification of Task Validity from Two Perpectives}
\textbf{(1) Perspective from Underlying TKG.} We take a commonly used temporal KB, i.e., ICEWS, as the KB for constructing underlying TKG ICEWS21. ICEWS-based TKGs contain socio-political facts. It is meaningful to perform forecasting over them because this can help to improve early warning in critical socio-political situations around the globe. \cite{DBLP:conf/icml/TrivediDWS17} has shown with case studies that ICEWS-based TKG datasets have underlying cause-and-effect temporal patterns and TKG forecasting models are built to capture them. This indicates that performing TKG forecasting over ICEWS-based TKGs are also valid. And therefore, developing forecasting TKGQA on top of ICEWS21 is meaningful and valid.
\textbf{(2) Perspective from the Motivation of Proposing Different Types of Questions.}
The motivation of proposing EPQs is to introduce TKG link forecasting (future link prediction) into KGQA, while
proposing YUQs is to introduce quadruple classification (stemming from triple classification) and yes-no type questions. 
We view quadruple classification in the forecasting scenario as deciding if the unseen TKG facts are valid based on previous known TKG facts. To answer EPQs and YUQs, models can be considered as understanding natural language questions first and then perform TKG reasoning tasks. Since TKG reasoning tasks are considered solvable and widely studied in the TKG community, our task over EPQs and YUQs is valid. 
We propose FRQs aiming to study the difference between humans and machines in fact reasoning. We have summarized the reasoning skills that are required to answer every FRQ in Fig. \ref{fig: mc example}, which also implies the potential direction for QA models to achieve improvement in fact reasoning in the future. We haven shown in Section \ref{sec: human} that our proposed FRQs are answerable to humans, which directly indicates the validity of our FRQs. Thus, answering FRQs in forecasting TKGQA is also valid and meaningful. 

\section{Conclusion}
In this work, we propose a novel task: forecasting TKGQA. To the best of our knowledge, it is the first work combining TKG forecasting with KGQA. We propose a coupled benchmark dataset F\textsc{orecast}TKGQ\textsc{uestions} that contains various types of questions including EPQs, YUQs and FRQs. To solve forecasting TKGQA, we propose F\textsc{orecast}TKGQA, a QA model that leverages a TKG forecasting model with a pre-trained LM. Though experimental results show that our model achieves great performance, there still exists a large room for improvement compared with humans. We hope our work can benefit future research and draw attention to studying the forecasting power of TKGQA methods.\\

\noindent
\textbf{Acknowledgement.} This work has been supported by the German Federal Ministry for Economic Affairs and Climate Action (BMWK) as part of the project CoyPu under grant number 01MK21007K.

\paragraph*{Supplemental Material Statement:}
Source code and data are uploaded here\footnote{https://github.com/ZifengDing/ForecastTKGQA}. Appendices are published in the arXiv version\footnote{https://arxiv.org/abs/2208.06501}. We have referred to the corresponding parts in the main body. Please check accordingly.
\bibliographystyle{splncs04}
\bibliography{mybibliography}
\appendix
\section{Scoring Function Details}
\label{sec:appendix score function}
\subsection{Entity Prediction}
The detailed definition of the EPQs' scoring function is defined as
\begin{equation}
\small
    \label{eq: es detail}
    \begin{aligned}
        \phi_{\text{ep}}(e) &= \text{Re}\left(<\mathbf{h}'_{(s_q,t_q)},\mathbf{h}_q, \Bar{\mathbf{h}}'_{(e,t_q)}>\right)\\
              &= \text{Re}\left( \sum_{k=1}^d \mathbf{h}'_{(s_q,t_q)}(k) \cdot \mathbf{h}_q(k) \cdot \Bar{\mathbf{h}}'_{(e,t_q)}(k) \right)\\
              &= <\text{Re}\left(\mathbf{h}'_{(s_q,t_q)}\right), \text{Re}\left(\mathbf{h}_q\right), \text{Re}\left(\mathbf{h}'_{(e,t_q)}\right)>\\
              & \quad   + <\text{Re}\left(\mathbf{h}'_{(s_q,t_q)}\right), \text{Im}\left(\mathbf{h}_q\right), \text{Im}\left(\mathbf{h}'_{(e,t_q)}\right)>\\
              & \quad   + <\text{Im}\left(\mathbf{h}'_{(s_q,t_q)}\right), \text{Re}\left(\mathbf{h}_q\right), \text{Im}\left(\mathbf{h}'_{(e,t_q)}\right)>\\
              & \quad   - <\text{Im}\left(\mathbf{h}'_{(s_q,t_q)}\right), \text{Im}\left(\mathbf{h}_q\right), \text{Re}\left(\mathbf{h}'_{(e,t_q)}\right)>.
    \end{aligned}
\end{equation}
\text{Re} and \text{Im} denote taking the real part and the imaginary part of the complex vector, respectively. 
$\mathbf{h}'_{(s_q,t_q)}, \mathbf{h}_q, \mathbf{h}'_{(e,t_q)} \in \mathbb{C}^d$. $\mathbf{h}'_{(s_q,t_q)}(k)$ denotes the $k$th element of it (same for $\mathbf{h}_q$ and $\mathbf{h}'_{(e,t_q)}$). $<\mathbf{v}_1, \mathbf{v}_2, \mathbf{v}_3> = \sum_{k=1}^d \mathbf{v}_1(k) \cdot \mathbf{v}_2(k) \cdot \mathbf{v}_3(k)$ denotes the dot product of three $d$-dimensional complex vectors $\mathbf{v}_1, \mathbf{v}_2, \mathbf{v}_3 \in \mathbb{C}^d$.
\subsection{Yes-Unknown}
The detailed definition of the YUQs' scoring function is defined as
\begin{equation}
\resizebox{0.95\columnwidth}{!}{$
\small
    \label{eq: ys detail}
    \begin{aligned}
        \phi_{\text{yu}}(x) &= \text{Re}\left(<\mathbf{h}'_{(s_q,t_q)},\mathbf{h}_q, \Bar{\mathbf{h}}'_{(o_q,t_q)},  \mathbf{h}_x>\right)\\
              &= \text{Re}\left( \sum_{k=1}^d \mathbf{h}'_{(s_q,t_q)}(k) \cdot \mathbf{h}_q(k) \cdot \Bar{\mathbf{h}}'_{(o_q,t_q)}(k) \cdot \mathbf{h}_x(k) \right)\\
              &= <\text{Re}\left(\mathbf{h}'_{(s_q,t_q)}\right), \text{Re}\left(\mathbf{h}_q\right), \text{Re}\left(\mathbf{h}'_{(o_q,t_q)}\right), \text{Re}\left(\mathbf{h}_x\right)>\\
              & \quad   + <\text{Re}\left(\mathbf{h}'_{(s_q,t_q)}\right), \text{Im}\left(\mathbf{h}_q\right), \text{Im}\left(\mathbf{h}'_{(o_q,t_q)}\right), \text{Re}\left(\mathbf{h}_x\right)>\\
              & \quad   + <\text{Im}\left(\mathbf{h}'_{(s_q,t_q)}\right), \text{Re}\left(\mathbf{h}_q\right), \text{Im}\left(\mathbf{h}'_{(o_q,t_q)}\right), \text{Re}\left(\mathbf{h}_x\right)>\\
              & \quad   + <\text{Re}\left(\mathbf{h}'_{(s_q,t_q)}\right), \text{Re}\left(\mathbf{h}_q\right), \text{Im}\left(\mathbf{h}'_{(o_q,t_q)}\right), \text{Im}\left(\mathbf{h}_x\right)>\\
              & \quad   - <\text{Im}\left(\mathbf{h}'_{(s_q,t_q)}\right), \text{Im}\left(\mathbf{h}_q\right), \text{Re}\left(\mathbf{h}'_{(o_q,t_q)}\right), \text{Re}\left(\mathbf{h}_x\right)>\\
              & \quad   - <\text{Im}\left(\mathbf{h}'_{(s_q,t_q)}\right), \text{Re}\left(\mathbf{h}_q\right), \text{Re}\left(\mathbf{h}'_{(o_q,t_q)}\right), \text{Im}\left(\mathbf{h}_x\right)>\\
              & \quad   - <\text{Im}\left(\mathbf{h}'_{(s_q,t_q)}\right), \text{Im}\left(\mathbf{h}_q\right), \text{Im}\left(\mathbf{h}'_{(o_q,t_q)}\right), \text{Im}\left(\mathbf{h}_x\right)>\\
              & \quad   - <\text{Re}\left(\mathbf{h}'_{(s_q,t_q)}\right), \text{Im}\left(\mathbf{h}_q\right), \text{Re}\left(\mathbf{h}'_{(o_q,t_q)}\right), \text{Im}\left(\mathbf{h}_x\right)>.
    \end{aligned}
$}
\end{equation}
\subsection{Fact Reasoning}
The detailed definition of FRQs' scoring function is defined as
\begin{equation}
\resizebox{0.95\columnwidth}{!}{$
\small
    \label{eq: cs detail}
    \begin{aligned}
        \phi_{\text{fr}}(c) &= \text{Re}\left(<\mathbf{h}'_{(s_c,t_c)},\mathbf{h}^{c}_q, \Bar{\mathbf{h}}'_{(o_c,t_c)},  \mathbf{h}'_q>\right)\\
              &= \text{Re}\left( \sum_{k=1}^d \mathbf{h}'_{(s_c,t_c)}(k) \cdot \mathbf{h}^{c}_q(k) \cdot \Bar{\mathbf{h}}'_{(o_c,t_c)}(k) \cdot \mathbf{h}'_q(k) \right)\\
              &= <\text{Re}\left(\mathbf{h}'_{(s_c,t_c)}\right), \text{Re}\left(\mathbf{h}^{c}_q\right), \text{Re}\left(\mathbf{h}'_{(o_c,t_c)}\right), \text{Re}\left(\mathbf{h}'_q(k)\right)>\\
              & \quad   + <\text{Re}\left(\mathbf{h}'_{(s_c,t_c)}\right), \text{Im}\left(\mathbf{h}^{c}_q\right), \text{Im}\left(\mathbf{h}'_{(o_c,t_c)}\right), \text{Re}\left(\mathbf{h}'_q(k)\right)>\\
              & \quad   + <\text{Im}\left(\mathbf{h}'_{(s_c,t_c)}\right), \text{Re}\left(\mathbf{h}^{c}_q\right), \text{Im}\left(\mathbf{h}'_{(o_c,t_c)}\right), \text{Re}\left(\mathbf{h}'_q(k)\right)>\\
              & \quad   + <\text{Re}\left(\mathbf{h}'_{(s_c,t_c)}\right), \text{Re}\left(\mathbf{h}^{c}_q\right), \text{Im}\left(\mathbf{h}'_{(o_c,t_c)}\right), \text{Im}\left(\mathbf{h}'_q(k)\right)>\\
              & \quad   - <\text{Im}\left(\mathbf{h}'_{(s_c,t_c)}\right), \text{Im}\left(\mathbf{h}^{c}_q\right), \text{Re}\left(\mathbf{h}'_{(o_c,t_c)}\right), \text{Re}\left(\mathbf{h}'_q(k)\right)>\\
              & \quad   - <\text{Im}\left(\mathbf{h}'_{(s_c,t_c)}\right), \text{Re}\left(\mathbf{h}^{c}_q\right), \text{Re}\left(\mathbf{h}'_{(o_c,t_c)}\right), \text{Im}\left(\mathbf{h}'_q(k)\right)>\\
              & \quad   - <\text{Im}\left(\mathbf{h}'_{(s_c,t_c)}\right), \text{Im}\left(\mathbf{h}^{c}_q\right), \text{Im}\left(\mathbf{h}'_{(o_c,t_c)}\right), \text{Im}\left(\mathbf{h}'_q(k)\right)>\\
              & \quad   - <\text{Re}\left(\mathbf{h}'_{(s_c,t_c)}\right), \text{Im}\left(\mathbf{h}^{c}_q\right), \text{Re}\left(\mathbf{h}'_{(o_c,t_c)}\right), \text{Im}\left(\mathbf{h}'_q(k)\right)>.
    \end{aligned}
$}
\end{equation}

\section{Implementation Details}
We implement all the experiments with PyTorch \cite{DBLP:conf/nips/PaszkeGMLBCKLGA19} on an NVIDIA A40 with 48GB memory and a 2.6GHZ AMD EPYC 7513 32-Core Processor. 
\subsection{TKG Forecasting} 
\label{sec:appendix tkg forecasting}
We train TANGO and TComplEx to perform TKG forecasting on ICEWS21. We implement TANGO with the official implementation\footnote{https://github.com/TemporalKGTeam/TANGO}. We switch its scoring function to ComplEx and perform a grid search for the embedding size (the dimension $d$ of the entity and relation representations). We keep the rest hyperparameters as TANGO's default setting of the ICEWS05-15 dataset. We train TComplEx with the official implementation\footnote{https://github.com/facebookresearch/tkbc}. We perform a grid search for the embedding size and keep the other hyperparameters as their default values. Table \ref{tab: embsize search space} provides the searching spaces of the grid searches for both methods. For each method, we run TKG forecasting experiments with different embedding sizes and choose the setting that leads to the best validation MRR as the best hyperparameter setting. We further run TANGO + TuckER with the best hyperparameters searched with TANGO + ComplEx for studying the effectiveness of different KG representations.
\begin{table}[htbp]
    \caption{Embedding size search space of TANGO and TComplEx. The embedding sizes leading to the best validation results are marked as bold. Note that the numbers represent the dimensions of complex space. Dimensions of real valued vectors are doubled, e.g., a complex vector with embedding size 100 will be transformed into a real valued vector with embedding size 200. The embedding size search spaces are taken from the default search space stated in the original papers of TANGO and TComplEx.}
    \label{tab: embsize search space}
    \centering
    \begin{tabular}{lc}
            \toprule 
        & \textbf{Emebdding Size Search Space} \\
       \midrule 
       TANGO & \{50, \textbf{100}, 150\}     \\
       TComplEx & \{\textbf{100}, 136, 174\}     \\
      \bottomrule 
        \end{tabular}
\end{table}

Besides, we train ComplEx on ICEWS21 for TKG forecasting. We use the implementation provided in the repository of TComplEx. Since ComplEx is not designed for processing temporal information, we transform every quadruple $(s,r,o,t)$ into a corresponding triplet $(s,r,o)$. We do not remove the repeated triplet. For example, if $(s,r,o,t_1)$ and $(s,r,o,t_2)$ both exist in the training set of ICEWS21, we train ComplEx with two identical triplets $(s,r,o)$. This preserves the inductive bias brought by the temporal knowledge base. To achieve a fairer comparison between ComplEx and TComplEx, we set the embedding size of ComplEx to 100 (same as the embedding size of TComplEx).

We report in Table \ref{tab: tkg val results} the validation results of the trained TKG models of all three KG reasoning methods on ICEWS21. We observe that TComplEx underperforms ComplEx in TKG forecasting. We attribute this to the excessive noise introduced by TComplEx's representations of unseen timestamps. Note that TComplEx is a TKG completion method. The validation timestamps are unseen during training, thus causing TComplEx to leverage the untrained timestamp representations during evaluation. ComplEx does not consider temporal information, which enables it to avoid the negative influence of the timestamps unseen in the training set. TANGO is designed for TKG forecasting. It outperforms the other methods greatly. Although TANGO + TuckER performs better than TANGO + ComplEx on ICEWS21, we choose the latter one for forecasting TKGQA since it aligns to our QA scoring function better (see Appendix \ref{sec:appendix KG representation} for detailed discussion).
\begin{table}
\caption{Validation results of KG reasoning models for TKG forecasting on ICEWS21.}
\label{tab: tkg val results}
\centering
    \resizebox{0.6\columnwidth}{!}{
\begin{tabular}{c c c c c } 
\toprule
\textbf{Metrics}& \textbf{MRR} & \textbf{Hits@1} & \textbf{Hits@3} & \textbf{Hits@10}\\ 
\midrule
ComplEx  & 0.278 & 0.188 & 0.312 & 0.456 \\ 
TComplEx  & 0.250 & 0.164 & 0.279 & 0.420 \\
TANGO + TuckER & 0.402 & 0.327 & 0.431 & 0.546 \\
TANGO + ComplEx & 0.389 & 0.324 & 0.411 & 0.515 \\ 
\bottomrule
\end{tabular}
}
\end{table}

\subsection{Baseline Details}
\label{sec:appendix baseline}
We use the library HuggingFace's Transformers \cite{DBLP:journals/corr/abs-1910-03771} to implement the pre-trained LMs, i.e., BERT and RoBERTa. Following C\textsc{ron}KGQA and TempoQR, we choose DistilBERT \cite{DBLP:journals/corr/abs-1910-01108} as the BERT model used throughout our work to save computational budget. For every natural language input, e.g., a natural language question, we take the output representation of the [CLS] token computed by an LM as its LM encoded representation.
\subsubsection{Pre-trained LM baselines for TKGQA} We provide detailed information of our pre-trained LM baselines. For EPQs, BERT and RoBERTa compute the scores of all entities with a prediction head $f_\text{ep}^\text{lm}: \mathbb{R}^{2d} \rightarrow \mathbb{R}^{|\mathcal{E}|}$ as
\begin{equation}
\label{eq: lm ep}
    \Phi_{\text{ep}} = f_\text{ep}^\text{lm}\left(\mathbf{h}_q\right).
\end{equation}
$\Phi_{\text{ep}}$ is a $|\mathcal{E}|$-dimensional real valued vector where each element corresponds to the score of an entity. $\mathbf{h}_q$ is the question representation output by BERT or RoBERTa with a projection to a $2d$ real space. Note that in F\textsc{orecast}TKGQA, we further map the $2d$ real valued vector to a $d$-dimensional complex vector. This step does not exist when we implement pre-trained LM baselines without including any TKG representation. We choose the entity with the highest score as the predicted answer. BERT\_int and RoBERTa\_int compute the score of each entity $e$ with a prediction head $f_\text{ep}^\text{lm\_int}: \mathbb{R}^{8d} \rightarrow \mathbb{R}^{1}$ as
\begin{equation}
\label{eq: lm int ep}
    \phi_{\text{ep}}(e) = f_\text{ep}^\text{lm\_int}\left(\mathbf{h}_{s}\|\mathbf{h}_q\|\mathbf{h}_{e}\|\mathbf{h}_{t_q}\right),
\end{equation}
where $\mathbf{h}_{s}$, $\mathbf{h}_{t_q}$, and $\mathbf{h}_{o}$ denote the TComplEx representations of the question's subject entity, the question's timestamp, and the entity $e$, respectively. Similarly, BERT\_ext and RoBERTa\_ext compute the score of each entity $e$ with a prediction head $f_\text{ep}^\text{lm\_ext}: \mathbb{R}^{6d} \rightarrow \mathbb{R}^{1}$ as
\begin{equation}
\label{eq: lm ext ep}
    \phi_{\text{ep}}(e) = f_\text{ep}^\text{lm\_ext}(\mathbf{h}_{(s_q,t_q)}\|\mathbf{h}_q\|\mathbf{h}_{(e,t_q)}),\\
\end{equation}
where $\mathbf{h}_{(s_q,t_q)}$ and $\mathbf{h}_{(e,t_q)}$ denote the TANGO representations of the question's subject entity and the entity $e$, respectively. Since TANGO and TComplEx representations are complex vectors in $\mathbb{C}^d$, we expand them into $2d$ real valued vectors, where the first half of every real valued vector is the real part of the original vector and the second half is the imaginary part. This applies to all the TKG representations used in pre-trained LM baselines for answering all three types of questions.

For yes-unknown questions, BERT and RoBERTa compute the scores of \textit{yes} and \textit{unknown} with a prediction head $f_\text{yu}^\text{lm}: \mathbb{R}^{2d} \rightarrow \mathbb{R}^{2}$ as
\begin{equation}
\label{eq: lm yn}
\Phi_{\text{yu}} = f_\text{yu}^\text{lm}(\mathbf{h}_q).
\end{equation}
$\Phi_{\text{yu}}$ is a 2-dimensional real valued vector where each element corresponds to the score of either \textit{yes} or \textit{unknown}. BERT\_int and RoBERTa\_int compute the score of each $x\in \{\textit{yes}, \textit{unknown}\}$ with a prediction head $f_\text{yn}^\text{lm\_int}: \mathbb{R}^{8d} \rightarrow \mathbb{R}^{1}$ as
\begin{equation}
\label{eq: lm int yn}
    \phi_{\text{yu}}(x) = f_\text{yn}^\text{lm\_int}\left(\mathbf{h}_{s_q}\|\mathbf{h}_q\|\mathbf{h}_{o_q}\|\mathbf{h}_{t_q}\right).
\end{equation}
And BERT\_ext and RoBERTa\_ext compute the score of each $x\in \{\textit{yes}, \textit{unknown}\}$ with a prediction head $f_\text{yu}^\text{lm\_ext}: \mathbb{R}^{6d} \rightarrow \mathbb{R}^{1}$ as
\begin{equation}
\label{eq: lm ext yn}
    \phi_{\text{yu}}(x) = f_\text{yu}^\text{lm\_ext}\left(\mathbf{h}_{(s_q,t_q)}\|\mathbf{h}_q\|\mathbf{h}_{(o_q,t_q)}\right).\\
\end{equation}
We choose the one (either \textit{yes} or \textit{unknown}) with the higher score as the predicted answer.

For every fact reasoning question, BERT and RoBERTa compute the score of the choice $c$ as
\begin{equation}
\label{eq: lm fr}
\phi_{\text{fr}}(c) = f_\text{fr}^\text{lm}(\mathbf{h}_q^{c}).
\end{equation}
$\mathbf{h}_q^{c}$ is the output of a pre-trained LM when the concatenation of the question $q$ and the choice $c$ is given as the input. $f_\text{fr}^\text{lm}: \mathbb{R}^{2d} \rightarrow \mathbb{R}^{1}$ is a layer of neural network for score computation. BERT\_int and RoBERTa\_int compute the score of the choice $c$ as
\begin{equation}
\label{eq: lm int fr}
\phi_{\text{fr}}(c) = f_\text{fr}^\text{lm\_int}\left(\mathbf{h}_q^\text{lm\_int}\|
                                     \mathbf{h}_{c}^\text{lm\_int}\right).
\end{equation}
$\mathbf{h}_q^\text{lm\_int} = \mathbf{h}_{s_q}\|\mathbf{h}_q^{c}\|\mathbf{h}_{o_q}\|\mathbf{h}_{t_q}$, where $\mathbf{h}_{s_q}$, $\mathbf{h}_{o_q}$ and $\mathbf{h}_{t_q}$ denote the TComplEx representations of the question's subject entity, object entity and timestamp, respectively. $\mathbf{h}_{c}^\text{lm\_int} = \mathbf{h}_{s_c}\|\mathbf{h}_q^{c}\|\mathbf{h}_{o_c}\|\mathbf{h}_{t_c}$, where $\mathbf{h}_{s_c}$, $\mathbf{h}_{o_c}$ and $\mathbf{h}_{t_c}$ denote the TComplEx representations of the choice's subject entity, object entity and timestamp, respectively. $f_\text{fr}^\text{lm\_int}: \mathbb{R}^{16d} \rightarrow \mathbb{R}^{1}$ is a layer of neural network for score computation. Similarly, BERT\_ext and RoBERTa\_ext compute the score of the choice $c$ as
\begin{equation}
\label{eq: lm ext fr}
\phi_{\text{fr}}(c) = f_\text{fr}^\text{lm\_ext}(\mathbf{h}_q^\text{lm\_ext}\|
                                     \mathbf{h}_{c}^\text{lm\_ext}).
\end{equation}
$\mathbf{h}_q^\text{lm\_ext} = \mathbf{h}_{(s_q,t_q)}\|\mathbf{h}_q^{c}\|\mathbf{h}_{(o_q,t_q)}$, where $\mathbf{h}_{(s_q,t_q)}$ and $\mathbf{h}_{(o_q,t_q)}$ denote the time-aware TANGO representations of the question's subject entity and object entity, respectively. $\mathbf{h}_{c}^\text{lm\_ext} = \mathbf{h}_{(s_c,t_c)}\|\mathbf{h}_q^{c}\|\mathbf{h}_{(o_c,t_c)}$, where $\mathbf{h}_{(s_c,t_c)}$ and $\mathbf{h}_{(o_c,t_c)}$ denote the time-aware TANGO representations of the choice's subject entity and object entity, respectively. $f_\text{fr}^\text{lm\_ext}: \mathbb{R}^{12d} \rightarrow \mathbb{R}^{1}$ is a layer of neural network for score computation.

\subsubsection{KGQA \& TKGQA Baselines}
For EmbedKGQA, we use the trained ComplEx representations as its supporting KG information. For C\textsc{ron}KGQA and TempoQR, we use the trained TComplEx representations as their supporting TKG information. We use the EmbedKGQA and C\textsc{ron}KGQA implementation provided in the repository of C\textsc{ron}KGQA\footnote{https://github.com/apoorvumang/CronKGQA}. We use the official implementation of TempoQR\footnote{https://github.com/cmavro/TempoQR}. Since we annotate the timestamps for every entity prediction question in F\textsc{orecast}TKGQ\textsc{uestions}, we do not implement soft/hard supervision proposed in TempoQR. We skip the soft/hard supervision and keep everything else as same as the original implementation. We implement all the KGQA baselines with their default hyperparameter settings.
\begin{table}[htbp] 
    \caption{F\textsc{orecast}TKGQA hyperparameter searching strategy.}
    \label{tab: hyperparameter search}
\small
    \begin{center}
    \resizebox{0.5\columnwidth}{!}{
    \begin{tabular}{cc} 
      \toprule 
     Hyperparameter & Search Space \\
       \midrule 
       TKG Model & \{TuckER, ComplEx\} \\
       Language Model & \{DistilBERT, RoBERTa\} \\
       Dropout & \{0.2, 0.3, 0.5\}\\
       Batch Size & \{32, 64, 128, 256, 512\}\\
       
      \bottomrule 
    \end{tabular}
    }
    \end{center}
\end{table}
\begin{table}[htbp] 
    \caption{Best hyperparameter setting.}
    \label{tab: Best hyperparameter setting}
\small
    \begin{center}
      \resizebox{0.8\columnwidth}{!}{
    \begin{tabular}{lcccc} 
      \toprule 
     \multicolumn{1}{l}{\textbf{Question Type}} & \multicolumn{1}{c}{\textbf{Entity Prediction}} & \multicolumn{1}{c}{\textbf{Yes-Unknown}} &\multicolumn{1}{c}{\textbf{Fact Reasoning}} \\
      \midrule 
      \textbf{Hyperparameter} &   &  &   \\
       \midrule 
       TKG Model & ComplEx & ComplEx & ComplEx\\
       Language Model & DistilBERT & DistilBERT & DistilBERT\\
       Dropout & 0.3& 0.3& 0.3\\
       Batch Size & 512 & 256 & 256\\
       
      \bottomrule 
    \end{tabular} 
    }
    \end{center}
\end{table}
\begin{table*}[ht!]
    \caption{Experimental results of EPQs on the validation set. Evaluation metrics are MRR and Hits@1/10.}\label{tab: val results ep}
    \small
    \centering
    \begin{tabular}{@{}lccc ccc ccc@{}}
\toprule
         & \multicolumn{3}{c}{\textbf{MRR}} & \multicolumn{3}{c}{\textbf{Hits@1}} & \multicolumn{3}{c}{\textbf{Hits@10}}\\
\cmidrule(lr){2-4} \cmidrule(lr){5-7} \cmidrule(lr){8-10}
\textbf{Model} & Overall & 1-Hop & 2-Hop & Overall & 1-Hop & 2-Hop & Overall & 1-Hop & 2-Hop\\
\midrule
        F\textsc{orecast}TKGQA 
        & 0.297 & 0.342 & 0.192 
        & 0.206 & 0.247 & 0.111
        & 0.475 & 0.526 & 0.353
        \\
\bottomrule
    \end{tabular}
\end{table*}
\begin{table*}[t]
    \caption{Standard deviation of the results of EPQs on the test set.}\label{tab: std results ep}
    \small
    \centering
    \begin{tabular}{@{}lccc ccc ccc@{}}
\toprule
         & \multicolumn{3}{c}{\textbf{MRR}} & \multicolumn{3}{c}{\textbf{Hits@1}} & \multicolumn{3}{c}{\textbf{Hits@10}}\\
\cmidrule(lr){2-4} \cmidrule(lr){5-7} \cmidrule(lr){8-10}
\textbf{Model} & Overall & 1-Hop & 2-Hop & Overall & 1-Hop & 2-Hop & Overall & 1-Hop & 2-Hop\\
\midrule
        F\textsc{orecast}TKGQA 
        & 0.0004 & 0.0004& 0.0009 
        & 0.0006 & 0.0007 & 0.0007
        & 0.0008 & 0.0008 & 0.0018
        \\
\bottomrule
    \end{tabular}
\end{table*}
\begin{table}[ht!]
    \caption{Experimental results of YUQs and FRQs on the validation sets. The evaluation metric is accuracy.}\label{tab: val results yn mc}
\small
    \centering
    \begin{tabular}{@{}lcc@{}}
\toprule
     & \multicolumn{2}{c}{\textbf{Accuracy}}\\
\cmidrule(lr){2-3}
\textbf{Question Type} & Yes-Unknown & Fact Reasoning\\
\midrule
        F\textsc{orecast}TKGQA & 0.873 & 0.758
         \\
\bottomrule
    \end{tabular}
\end{table}
\begin{table}[ht!]
\caption{Standard deviation of the results of YUQs and FRQs on the test set.}\label{tab: std results yn mc}
\small
    \centering
    \begin{tabular}{@{}lcc@{}}
\toprule
     & \multicolumn{2}{c}{\textbf{Accuracy}}\\
\cmidrule(lr){2-3}
\textbf{Question Type} & Yes-Unknown & Fact Reasoning\\
\midrule
        F\textsc{orecast}TKGQA & 0.0013 & 0.0052
         \\
\bottomrule
    \end{tabular}
\end{table}
\begin{table}[ht!] 
    \caption{GPU memory usage.}
    \label{tab: memory}
\small
    \begin{center}
      \resizebox{0.8\columnwidth}{!}{
    \large\begin{tabular}{lcccc} 
      \toprule 
     \multicolumn{1}{l}{\textbf{Question Type}} & \multicolumn{1}{c}{\textbf{Entity Prediction}} & \multicolumn{1}{c}{\textbf{Yes-Unknown}} &\multicolumn{1}{c}{\textbf{Fact Reasoning}} \\
      \midrule 
      \textbf{Model} & GPU Memory  & GPU Memory & GPU Memory  \\
       \midrule 
       F\textsc{orecast}TKGQA & 45,239MB & 12,241MB & 22,719MB  \\
       
      \bottomrule 
    \end{tabular} 
    }
    \end{center}
\end{table}
\begin{table}[ht!]
     \caption{Training time (second) of F\textsc{orecast}TKGQA on all types of questions.}\label{tab: train time}
\small
    \centering
    \resizebox{0.8\columnwidth}{!}{
    \large\begin{tabular}{@{}lccc@{}}
\toprule
        \textbf{Question Type} & \multicolumn{1}{c}{\textbf{Entity Prediction}} & \multicolumn{1}{c}{\textbf{Yes-Unknown}} & \multicolumn{1}{c}{\textbf{Fact Reasoning}}\\
\midrule
        \textbf{Model} & & &\\
\midrule
        F\textsc{orecast}TKGQA  & 63,840 & 3,700 & 5000
       \\
\bottomrule
    \end{tabular}
    }
\end{table}
\begin{table}[ht!]
    \caption{Test time (second) of F\textsc{orecast}TKGQA on all types of questions.}\label{tab: test time}
\small
    \centering
    \resizebox{0.8\columnwidth}{!}{
    \large\begin{tabular}{@{}lccc@{}}
\toprule
        \textbf{Question Type} & \multicolumn{1}{c}{\textbf{Entity Prediction}} & \multicolumn{1}{c}{\textbf{Yes-Unknown}} & \multicolumn{1}{c}{\textbf{Fact Reasoning}}\\
\midrule
        \textbf{Model} & & &\\
\midrule
        F\textsc{orecast}TKGQA  & 48 & 33 & 3
       \\
\bottomrule
    \end{tabular}
    }
\end{table}
\begin{table}[ht!]
 \caption{Number of parameters of F\textsc{orecast}TKGQA on all types of questions.}\label{tab: num parameter}
\small
    \centering
    \resizebox{0.8\columnwidth}{!}{
    \large\begin{tabular}{@{}lccc@{}}
\toprule
        \textbf{Question Type} & \multicolumn{1}{c}{\textbf{Entity Prediction}} & \multicolumn{1}{c}{\textbf{Yes-Unknown}} & \multicolumn{1}{c}{\textbf{Fact Reasoning}}\\
\midrule
        \textbf{Model} & & &\\
\midrule
        F\textsc{orecast}TKGQA  & 234,600 & 234,600 & 354,800
       \\
\bottomrule
    \end{tabular}
    }
\end{table}
\subsection{F\textsc{orecast}TKGQA}
We search hyperparameters of F\textsc{orecast}TKGQA following Table \ref{tab: hyperparameter search}. For every type of question, we do 60 trials, and let our model run for 50 epochs. We select the trial leading to the best performance on the validation set and take this hyperparameter setting as our best configuration. We train our model five times with different random seeds and report averaged results. The best hyperparameters concerning all three types of questions are shown in Table \ref{tab: Best hyperparameter setting}. We also report the model performance on the validation sets in Table \ref{tab: val results ep} and Table \ref{tab: val results yn mc}. We further report the standard deviation of the results on the test sets in Table \ref{tab: std results ep} and Table \ref{tab: std results yn mc}. The GPU memory usage is reported in Table \ref{tab: memory}. The training time and test time of our model are presented in Table \ref{tab: train time} and Table \ref{tab: test time}. The number of parameters of our model is presented in Table \ref{tab: num parameter}.

\section{F\textsc{orecast}TKGQ\textsc{uestions} Details}
\subsection{Natural Language Relation Template}
After we get ICEWS21, we get a TKG with 253 relation types. 
We create natural language relation templates for 250 out of 253 relation types for question generation.
The rest three relation types in ICEWS21 are not taken into consideration because either the verb is not suited for a question in the future tense (\textit{Attempt to assassinate}) or there is no clear description for the subject-object-relationship of the relation type in \cite{DVN/28075_2015} (\textit{Demobilize armed forces} and \textit{Demonstrate military or police power}). We use the generated relation templates for question generation of all three types of questions. For fact reasoning questions, we also use these relation templates to generate natural language choices.
\subsection{Natural Language Question Template}
All question templates are presented in \textit{Question\_Generation/template\_icews.xlsx} which is attached with the submission in Easychair. 2-hop EPQs and their templates are generated with  \textit{Question\_Generation/generate\_qa\_anyburl.py}.
\subsection{2-Hop EPQ Generation Details}
\label{sec:appendix 2hop EPQ generation}
We generate 2-hop questions by utilizing AnyBURL \cite{https://doi.org/10.48550/arxiv.2004.04412}, a rule-based KG reasoning model. We first split ICEWS21 into TKG snapshots, where each snapshot $\mathcal{G}_{t_i} = \{(s,r,o,t)\in \mathcal{G}|t = t_i\}$ contains all the TKG facts happening at the same timestamp. We treat every TKG snapshot as a non-temporal KG and train an AnyBURL model with the KG completion task on each TKG snapshot for rule extraction (KG completion aims to predict the missing entity from every query $(s,r,?)$). Since AnyBURL is a KG reasoning method that cannot process temporal information, we transform every quadruple $(s,r,o,t)$ into a corresponding triplet $(s,r,o)$. For each TKG snapshot, we keep the 2-hop rules with a confidence higher than 0.5 extracted by AnyBURL, and manually check if two associated TKG facts in each rule potentially have a logical causation or can be used to interpret positive/negative entity relationships. After this process, we take the remaining 2-hop rules as the drafts for generating 2-hop EPQ templates. The complete list of extracted 2-hop rules is presented in \textit{Question\_Generation/anyburl\_ICEWS.txt}. 2-hop EPQs and their templates are generated with \textit{Question\_Generation/generate\_qa\_anyburl.py}, given the extracted rules.
\subsection{FRQ Generation Details}
\label{sec:appendix FRQ generation}
We train xERTE \cite{DBLP:conf/iclr/HanCMT21} on ICEWS21 for TKG forecasting, and pick out all the link prediction queries $(s,r,?,t)$ whose ground-truth missing entities are ranked by xERTE as top 1. We collect the TKG facts corresponding to these queries for question generation. The intuition of this step is that we assume that the better xERTE performs on a link prediction query, the more reasonable the returned prior facts are for explainability. Ranking the ground-truth missing entities as top 1 indicates that xERTE performs very well on these link prediction queries. We wish to use xERTE to generate reasonable fact reasoning questions, therefore, we want it to find reasonable supporting evidence of the TKG facts by returning relevant prior facts. For each collected top 1 fact, we take the prior facts with the highest contribution score, the lowest contribution score, the median contribution score, and the second highest contribution score as the facts for generating the choices \textbf{Answer}, \textbf{Negative}, \textbf{Median} and \textbf{Hard Negative}, respectively. In this way, we can generate a large number of question candidates by fitting the corresponding facts into question templates.

After we collect all the question candidates, we have 78,606 questions. We find that there exist a large number of question candidates whose question and \textbf{Answer} share the same $s,r,o$. For example, the TKG fact of a question candidate is (\textit{Sudan}, \textit{host}, \textit{Ramtane Lamamra}, \textit{2021-08-01}), and the TKG fact of its \textbf{Answer} is (\textit{Sudan}, \textit{host}, \textit{Ramtane Lamamra}, \textit{2021-07-29}). We filter out all the question candidates with this pattern since we think that they are not satisfying our motivation for proposing fact reasoning questions. We wish to generate the questions that require fact reasoning, rather than finding the repeated facts happening at different timestamps. A good example of the questions we want to generate is as follows. For the question whose associated fact is (\textit{Envoy (United States)}, \textit{visit}, \textit{China}, \textit{2021-08-31}), the associated fact of its \textbf{Answer} is (\textit{Envoy (United States)}, \textit{express the intent to meet or negotiate}, \textit{China}, \textit{2021-08-30}). From human knowledge, \textbf{Answer}'s fact serves as a highly possible reason for the fact in the question, and it is also diverse from the question fact. To this end, we have 50,379 question candidates left.

\begin{figure*}
    \centering
    \includegraphics[width=\textwidth]{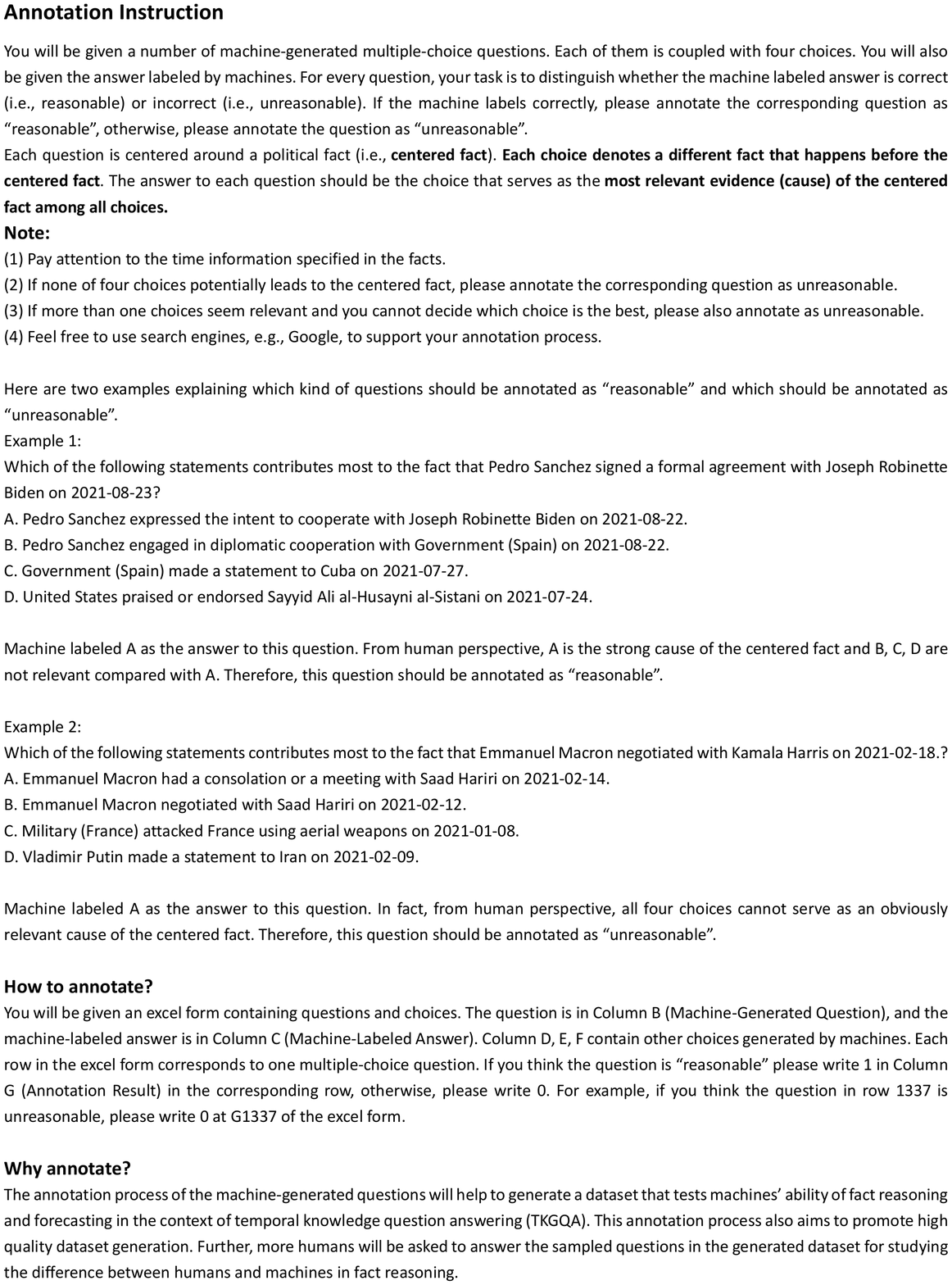}
    \caption{Human annotation instruction for fact reasoning questions.}
    \label{fig:human_annotation_instruction}
\end{figure*}

\begin{figure*}
    \centering
    \includegraphics[width=\textwidth]{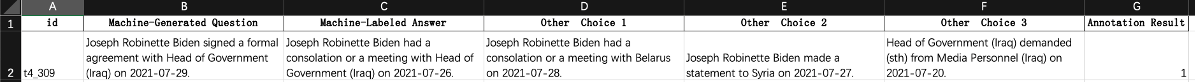}
    \caption{Human annotation interface for fact reasoning questions.}
    \label{fig:human_annotation_interface}
\end{figure*}

We then ask five graduate students (major in computer science) to further annotate the remaining question candidates by deciding whether each of them is reasonable or not. Students are allowed to use their own knowledge and search engines to help annotation. If the students think that a question's \textbf{Answer} is not the most contributive to the question, they are asked to annotate this question as unreasonable, otherwise, they are asked to annotate it as reasonable. For every question candidate, if it is annotated as unreasonable by three students, we filter it out. As a result, we have 4,195 questions left. We use Fleiss' kappa to measure inter-annotator agreement. Fleiss' kappa is 0.63 in our annotation process. The estimated annotation time for each student is 320 hours. The annotation instruction and interface are presented in Figure \ref{fig:human_annotation_instruction} and \ref{fig:human_annotation_interface}, respectively.
\begin{figure}[htbp]
\centering

\subfloat[EPQs and YUQs.]{%
  \includegraphics[width=0.5\columnwidth]{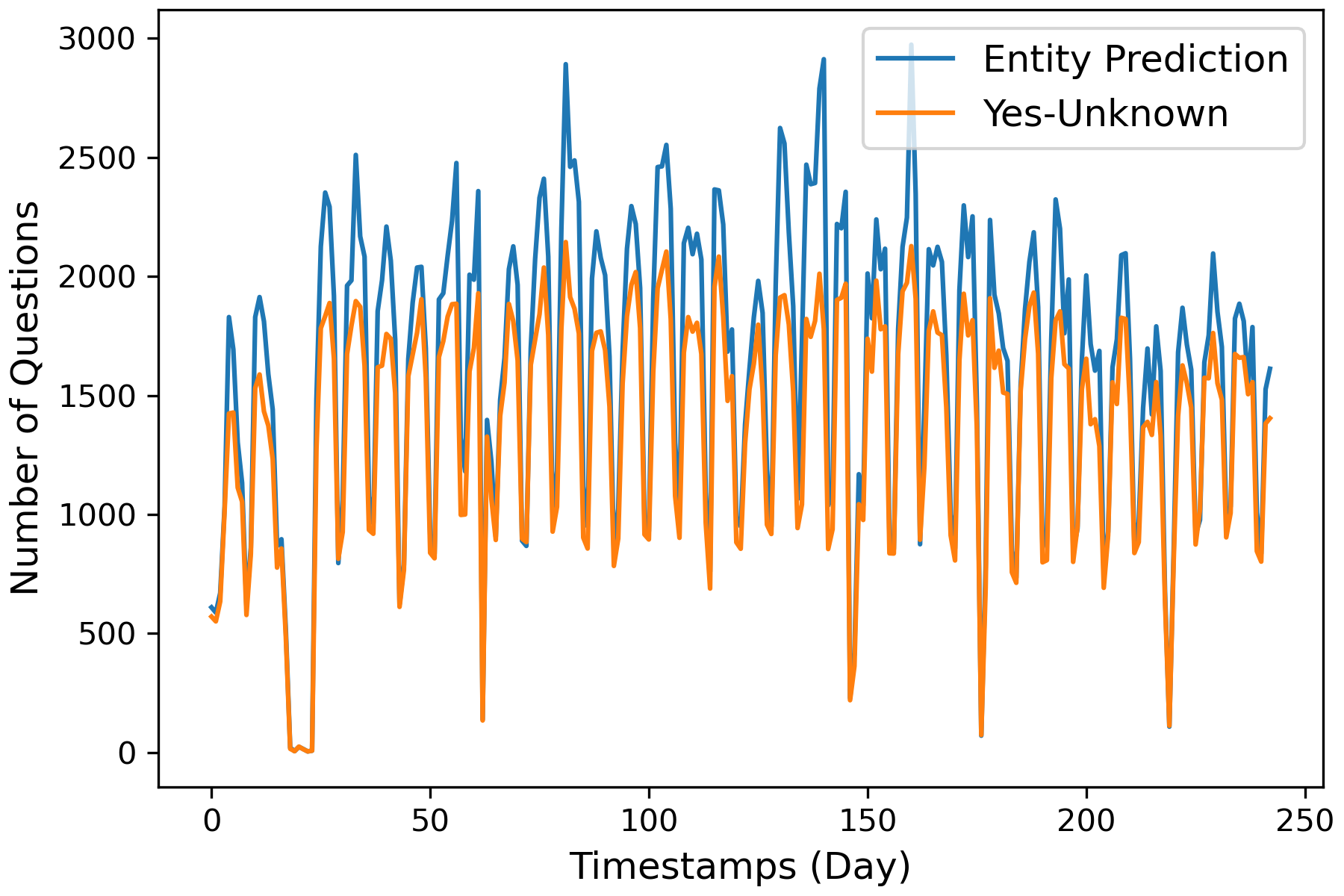}%
  \label{fig: ep yu distribution}%
}\hfill
\subfloat[FRQs.]{%
  \includegraphics[width=0.5\columnwidth]{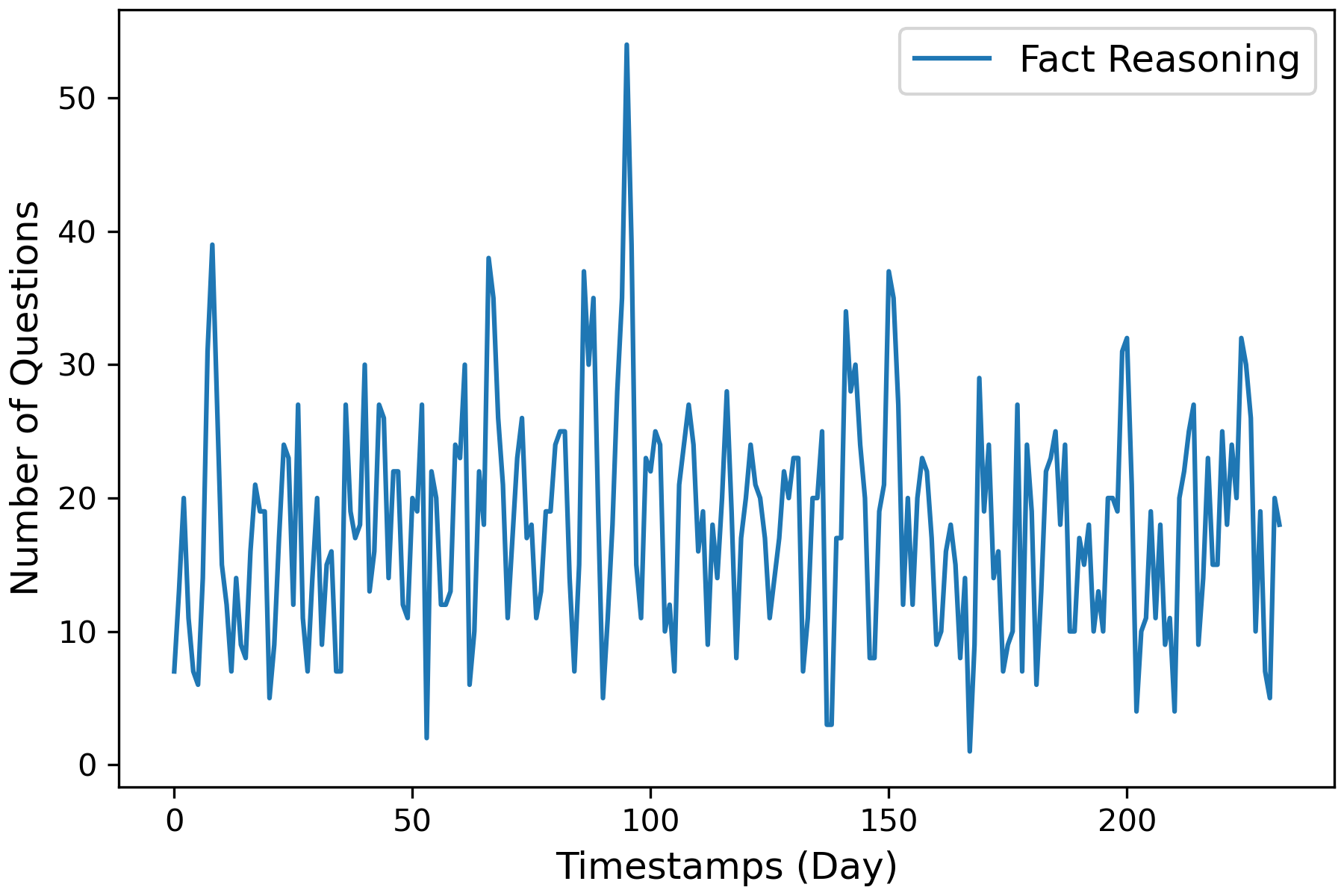}%
  \label{fig: fr distribution}%
}

\caption{Question distribution of different types of questions along the time axis.}
\end{figure}
\subsection{Question Time Distribution}
We provide the distribution of the questions along the time axis of our dataset in Figure \ref{fig: ep yu distribution} and Figure \ref{fig: fr distribution}. We plot the number of questions at every timestamp for all three types of questions. The numbers on the horizontal axis denote how many days away from 2021-01-01.

\section{Full Experimental Results on EPQs}
\label{sec:appendix full exp epq}
We present Table \ref{tab: main results ep full} as the supplement of the main results regarding EPQs in the main paper. We present the aggregated overall performance of MRR and Hits@k.
\begin{table*}[htbp]
    \caption{Complete experimental results of EPQs. The best results are marked in bold.}\label{tab: main results ep full}
    \small
    \centering
    \begin{tabular}{@{}lccc ccc ccc@{}}
\toprule
         & \multicolumn{3}{c}{\textbf{MRR}} & \multicolumn{3}{c}{\textbf{Hits@1}} & \multicolumn{3}{c}{\textbf{Hits@10}}\\
\cmidrule(lr){2-4} \cmidrule(lr){5-7} \cmidrule(lr){8-10}
\textbf{Model} & Overall & 1-Hop & 2-Hop & Overall & 1-Hop & 2-Hop & Overall & 1-Hop & 2-Hop\\

\midrule
        RoBERTa & 0.161 & 0.166 & 0.149 
        & 0.098 & 0.104 & 0.085 
        & 0.282 & 0.288 & 0.268 
         \\
        BERT 
        & 0.253	& 0.279	& 0.182	
        & 0.168	& 0.192	& 0.106	
        & 0.421	& 0.451	& 0.342
         \\
\midrule
        EmbedKGQA 
        & 0.278	& 0.317	& 0.185	
        & 0.194	& 0.228	& 0.112	
        & 0.443	& 0.489	& 0.333
        \\
\midrule 
        RoBERTa\_int
        & 0.246	& 0.283	& 0.157	
        & 0.162	& 0.190	& 0.094	
        & 0.415	& 0.467	& 0.290
         \\
        BERT\_int
        & 0.275	& 0.314	& 0.183	
        & 0.189	& 0.223	& 0.107	
        & 0.447	& 0.490	& 0.344
        \\
        C\textsc{ron}KGQA 
        & 0.119	& 0.131	& 0.090	
        & 0.069	& 0.081	& 0.042	
        & 0.218	& 0.231	& 0.187
         \\
        TempoQR 
        & 0.134	& 0.145	& 0.107	
        & 0.085	& 0.094	& 0.061	
        & 0.230	& 0.243	& 0.199
         \\
\midrule
        RoBERTa\_ext 
        & 0.269	& 0.306	& 0.180	
        & 0.184	& 0.216	& 0.108	
        & 0.433	& 0.497	& 0.323
         \\
        BERT\_ext
        & 0.295	& 0.331	& 0.208	
        & 0.206	& 0.239	& 0.128	
        & 0.467	& 0.508	& 0.369
        \\    
\midrule
        F\textsc{orecast}TKGQA 
        & \textbf{0.303} & \textbf{0.339} & \textbf{0.216}
        & \textbf{0.213} & \textbf{0.248} & \textbf{0.129}
        & \textbf{0.478} & \textbf{0.517} & \textbf{0.386}
        \\
\bottomrule
    \end{tabular}
\end{table*}

\section{Human Benchmark Details}
\label{sec:appendix human benchmark}
\begin{figure}[htbp]
\centering
\subfloat[Survey insturction.]{%
  \includegraphics[width=0.85\columnwidth]{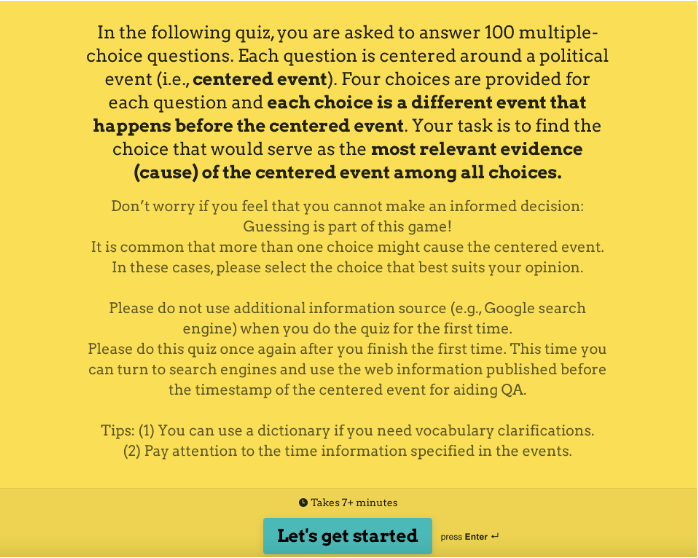}%
  \label{fig: survey instruction}%
}\qquad
\subfloat[Survey interface.]{%
  \includegraphics[width=0.85\columnwidth]{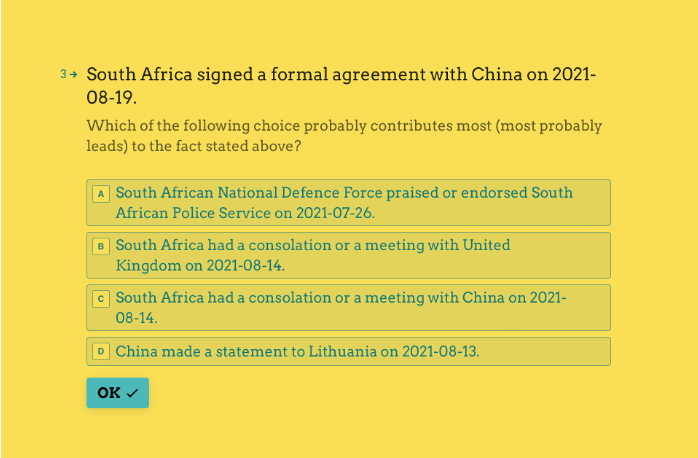}%
  \label{fig: survey interface}%
}
\caption{Human benchmark survey instruction and interface.}
\end{figure}

We ask five graduate students (major in computer science, not participating in annotation during FRQ generation) to answer 100 questions randomly sampled from the test set of FRQs. We consider two settings: (a) Humans answer FRQs with their own knowledge and inference ability. \textbf{Search engines are not allowed}; (b) Humans can turn to search engines and use the web information published \textbf{before the question timestamp} for aiding QA. We create a survey that contains the selected 100 questions. Figure \ref{fig: survey instruction} and \ref{fig: survey interface} show the instruction of survey and the interface of answering. We first ask the students to do the survey in setting (a), and then ask them to do it once again in setting (b). The ground-truth answers to survey questions are not shown to students throughout the whole process. Also, students have no idea which question they answer incorrectly. Thus, they cannot use this information to exclude wrong choices when they do the survey for the second time. From Table 5 of the main paper, we observe that with search engines, humans can better answer FRQs, although humans can already reach 0.936 accuracy without any additional information source.

\begin{figure}
    \centering
    \includegraphics[width=\columnwidth]{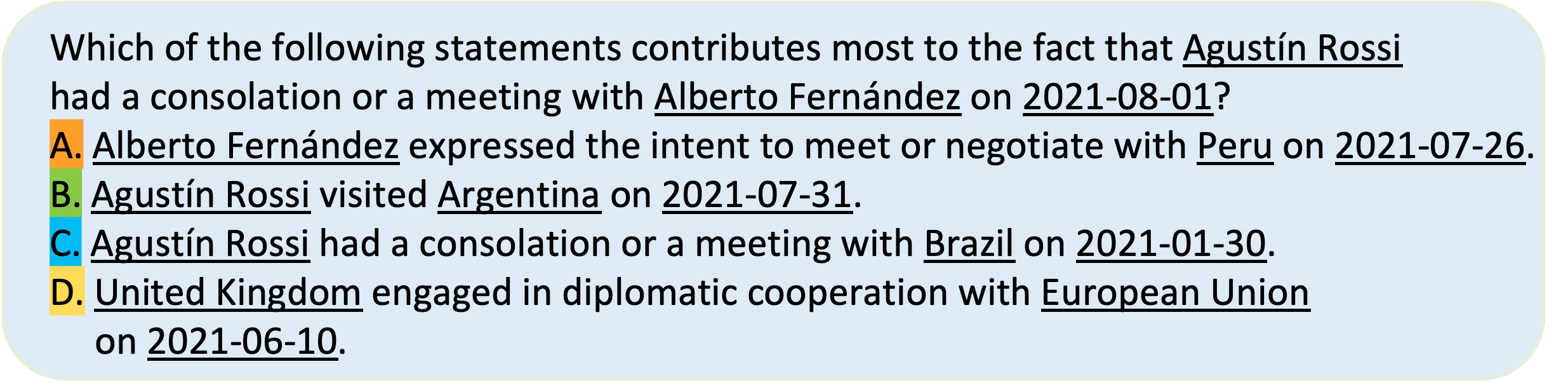}
    \caption{Example question in the human benchmark survey.}
    \label{fig:human example}
\end{figure}

\paragraph{Example to explain accuracy improvement from setting (a) to (b).} We present an example explaining the human performance improvement from setting (a) to (b). Figure \ref{fig:human example} shows a question in the generated survey for human benchmark. In setting (a), 3 of 5 students make a mistake by choosing A. After being allowed to use search engines in setting (b), they all choose the correct choice B. This is because in setting (a), most students have no idea that \textit{Alberto Fernández} is the president of Argentina. But after using search engines, they know the identity of \textit{Alberto Fernández} and manage to achieve correct reasoning.

\section{Details of Multi-Hop Scorer}
\label{sec:appendix mhs details}
\begin{figure*}[t]
    \centering
    \includegraphics[width=\textwidth]{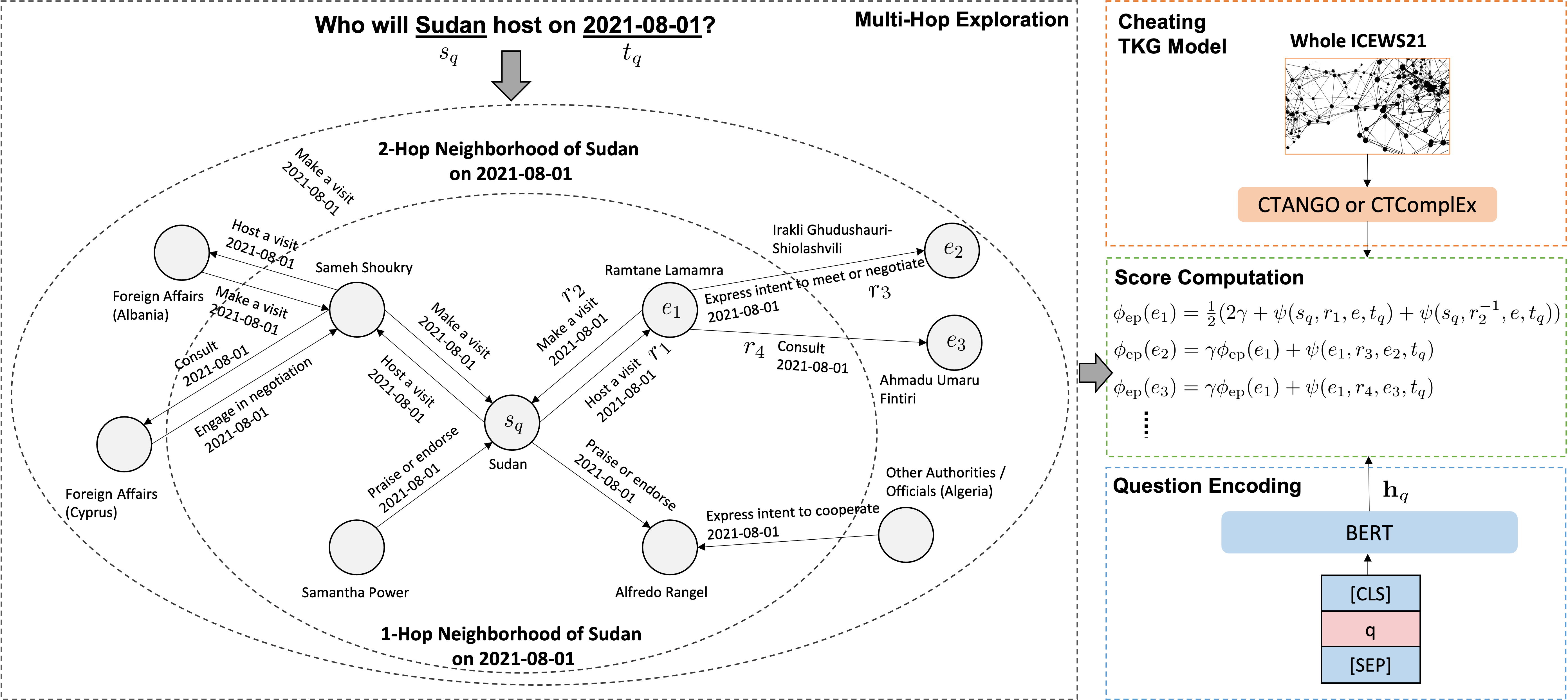}
    \caption{Assume we have a question: \textit{Who will \underline{Sudan} host on \underline{2021-08-01}?} The annotated subject entity $s_q$ is \textit{Sudan} and the annotated timestamp $t_q$ is \textit{2021-08-01}. We first pick the snapshot $\mathcal{G}_{t_q}$ and find $s_q$'s $n$-hop ($n$ = 2 in our case) neighbors on $\mathcal{G}_{t_q}$. Starting from $s_q$, MHS updates the scores of outer entities for 2-hops until all $s_q$'s 2-hop neighbors on $\mathcal{G}_{t_q}$ (\{\textit{Ramtane Lamamra} ($e_1$), \textit{Sameh Shoukry}, \textit{Samantha Power}, \textit{Alfredo Rangel}, \textit{Irakli Ghudushauri-Shiolashvili} ($e_2$), \textit{Ahmadu Umaru Fintiri} ($e_3$), \textit{Other Authorities/Officials (Algeria)}, \textit{Foreign Affairs (Albania)}, \textit{Foreign Affairs (Cyprus)} \} in our example) are visited. Initially, MHS assigns a score of 1 to $s_q$ and 0 to any other unvisited entity. To be specific, MHS first propagates scores to $s_q$'s 1-hop neighbors on $\mathcal{G}_{t_q}$, e.g., $e_1$. Then through the visited 1-hop neighbors, MHS propagates scores to $s_q$'s 2-hop neighbors. Score computation for $e_1$, $e_2$, $e_3$ is presented in this figure. $r_2^{-1}$ denotes the inverse relation of $r_2$ that points from $s_q$ to $e_1$. We transform $r_2$ to $r_2^{-1}$ because we define the 1-hop neighbor of an entity with its incoming edges (following TANGO \cite{DBLP:conf/emnlp/HanDMGT21}). Scores are computed by considering the graphical structure of $\mathcal{G}_{t_q}$. After the score propagation process, the entity with the highest score is taken as the predicted answer $e_\text{ans}$.}
    \label{fig:mhs structure}
\end{figure*}

We develop a QA model, i.e., Multi-Hop Scorer (MHS), for \textbf{non-forecasting TKGQA} (the TKGQA task proposed in \cite{DBLP:conf/acl/SaxenaCT20}). We use it to prove that given the ground-truth TKG information at the question timestamp $t_q$ (same setting as non-forecasting TKGQA), the EPQs in F\textsc{orecast}TKGQ\textsc{uestions} are answerable. 
Considering the non-forecasting setting, we equip MHS with two cheating TKG models (CTComplEx and CTANGO) and also design MHS by considering the multi-hop graphical structure of the snapshot $\mathcal{G}_{t_q} = \{(s,r,o,t)\in \mathcal{G}|t = t_q\}$. We illustrate MHS's model structure with an example in Fig. \ref{fig:mhs structure}. Starting from the annotated subject entity $s_q$ of an EPQ, MHS updates the scores of outer entities for $n$-hops ($n=2$ in our experiments) until all $s_q$'s $n$-hop neighbors on the snapshot $\mathcal{G}_{t_q}$ are visited. Initially, MHS assigns a score of 1 to $s_q$ and 0 to any other unvisited entity. 
For each unvisited entity $e$, it then computes $e$'s score as: 
\begin{equation}
\resizebox{0.6\columnwidth}{!}{$
\begin{aligned}
        &\Bar{\phi}_{\text{ep}}(e) = \sum_{(e',r)\in\mathcal{N}_e(t_q)}(\gamma \cdot \phi_{\text{ep}}(e') + \psi(e',r,e,t_q)), \\
        &\phi_{\text{ep}}(e) = \frac{1}{|\mathcal{N}_e(t_q)|} \Bar{\phi}_{\text{ep}}(e),
\end{aligned}
$}
\end{equation}
where $\mathcal{N}_e(t_q) = \{(e', r)|(e',r,e,t_q)\in \mathcal{G}_{t_q}\}$ is $e$'s 1-hop neighborhood on the snapshot $\mathcal{G}_{t_q}$ and $\gamma$ is a discount factor. We couple MHS with CTComplEx and CTANGO, and define $\psi(e',r,e,t_q)$ separately. For MHS + CTComplEx, we define
\begin{equation}
\resizebox{0.6\columnwidth}{!}{$
    \psi(e',r,e,t_q) = f_2(f_1(\mathbf{h}_{e'}\|\mathbf{h}_r\|\mathbf{h}_{e}\|\mathbf{h}_{t_q}\|\mathbf{h}_q)). 
$}
\end{equation} 
$f_1: \mathbb{R}^{10d}\rightarrow \mathbb{R}^{2d}$, $f_2: \mathbb{R}^{2d}\rightarrow \mathbb{R}^{1}$ are two neural networks.
$\mathbf{h}_{e}, \mathbf{h}_{e'}, \mathbf{h}_{r}, \mathbf{h}_{t_q}$ are the CTComplEx representations of entities $e$, $e'$, relation $r$ and timestamp $t_q$, respectively. 
For MHS + CTANGO, we take the idea of F\textsc{orecast}TKGQA and define 
\begin{equation}
\resizebox{0.6\columnwidth}{!}{$
    \psi(e',r,e,t_q) = \text{Re}\left(<\mathbf{h}_{(e',t_q)},\mathbf{h}_r, \Bar{\mathbf{h}}_{(e,t_q)}, \mathbf{h}_q>\right).
$}
\end{equation} 
$\mathbf{h}_{(e,t_q)}$, $\mathbf{h}_{(e',t_q)}$ are the CTANGO entity representations of $e$, $e'$ at $t_q$, respectively. $\mathbf{h}_r$ is the CTANGO relation representation of $r$. $\mathbf{h}_q$ is BERT encoded question representation.

\section{Further Analysis on F\textsc{orecast}TKGQA}
\label{sec:appendix KG representation}
\subsubsection{Ablation on KG Representations}
We conduct an ablation study by comparing the performance of F\textsc{orecast}TKGQA coupled with different KG (TKG) representations.
We first train ComplEx on ICEWS21 and provide our model with its representations. We observe in Table \ref{tab: kg represent} that TANGO representations are more effective than static KG representations in our proposed model. Besides, we switch TANGO's scoring function to TuckER \cite{DBLP:conf/emnlp/BalazevicAH19} when we train TANGO on ICEWS21. Table \ref{tab: kg represent} shows that TANGO + ComplEx aligns better to our QA module.
\begin{table*}[htbp]
    \caption{Comparison of different KG representations. w. means with.
    EPQ, YUQ, FRQ represent entity prediction, yes-unknown and fact reasoning questions, respectively.
    }\label{tab: kg represent}
\small
    \centering
    \resizebox{\textwidth}{!}{
    \begin{tabular}{@{}lccccccccc cc @{}}
\toprule
        \textbf{Question Type} & \multicolumn{9}{c}{\textbf{EPQ}} &\textbf{YUQ} & \textbf{FRQ}\\
\cmidrule(lr){2-10} 
&\multicolumn{3}{c}{\textbf{MRR}} & \multicolumn{3}{c}{\textbf{Hits@1}} & \multicolumn{3}{c}{\textbf{Hits@10}} & \textbf{Accuracy} & \textbf{Accuracy}\\
\cmidrule(lr){2-4} \cmidrule(lr){5-7} \cmidrule(lr){8-10} 
\textbf{Model} & Overall & 1-Hop & 2-Hop & Overall & 1-Hop & 2-Hop & Overall & 1-Hop & 2-Hop \\
\midrule
        F\textsc{orecast}TKGQA w. ComplEx 
        & 0.296	& 0.338	& 0.196	
        & 0.207	& 0.245	& 0.114	
        & 0.470	& 0.516	& 0.358
        & 0.863 & 0.752  
         \\
        F\textsc{orecast}TKGQA w. TANGO + TuckER 
        & 0.298	& 0.335	& 0.211	
        & 0.210	& 0.245	& 0.125	
        & 0.474	& 0.511	& 0.385
        & 0.867 & 0.757  
         \\
\midrule
        F\textsc{orecast}TKGQA w. TANGO + ComplEx 
        & \textbf{0.303} & \textbf{0.339} & \textbf{0.216}
        & \textbf{0.213} & \textbf{0.248} & \textbf{0.129}
        & \textbf{0.478} & \textbf{0.517} & \textbf{0.386}
        & \textbf{0.870} & \textbf{0.769}
        \\
\bottomrule
    \end{tabular}
     }

\end{table*}
\end{document}